\documentclass{article}

\PassOptionsToPackage{numbers, compress}{natbib}

\usepackage[final]{neurips_2022}

\usepackage[utf8]{inputenc} 
\usepackage[T1]{fontenc}    
\usepackage{url}            
\usepackage{booktabs}       
\usepackage{amsfonts}       
\usepackage{nicefrac}       
\usepackage{microtype}      
\usepackage[dvipsnames]{xcolor}       

\usepackage{amsthm}

\usepackage[colorlinks=true,linkcolor=Cerulean,citecolor=Cerulean]{hyperref}

\usepackage{researchpack}
\usepackage{mathtools}
\usepackage{xspace}
\usepackage{enumitem}
\usepackage{wrapfig}
\usepackage[capitalise,nameinlink]{cleveref}

\usepackage{tikz}
\usetikzlibrary{positioning,shapes,arrows,calc}

\usepackage{pifont}

\definecolor{cbnmcolor}{HTML}{0070ff}
\definecolor{glancecolor}{HTML}{ff0070}

\newcommand{\EM}[1]{}
\newcommand{\ST}[1]{}
\newcommand{\AP}[1]{}

\newcommand{\ie}{i.e.,\xspace}
\newcommand{\eg}{e.g.,\xspace}
\newcommand{\etc}{\textit{etc}.\xspace}

\newcommand{\defeq}{\ensuremath{:=}}
\newcommand{\softmax}{\ensuremath{\mathrm{softmax}}}

\newcommand{\indep}{\mathrel{\text{\scalebox{1.07}{$\perp\mkern-10mu\perp$}}}}

\newcommand{\acronym}{aliGned LeAk-proof coNCEptual Networks\xspace}
\newcommand{\method}{GlanceNet\xspace}
\newcommand{\methods}{GlanceNets\xspace}

\newcommand{\KL}{\ensuremath{\mathsf{KL}}}

\newcommand{\expl}{\ensuremath{\calE}\xspace}
\newcommand{\dataset}{\ensuremath{D}\xspace}

\title{\methods: Interpretable, Leak-proof\\ Concept-based Models}

\author{%
    Emanuele Marconato \\
    DISI \\
    University of Trento and University of Pisa \\
    Trento, Italy \\
    \texttt{emanuele.marconato@unitn.it} \\
    \And
    Andrea Passerini \\
    DISI \\
    University of Trento \\
    Trento, Italy \\
    \texttt{andrea.passerini@unitn.it} \\
    \And
    Stefano Teso \\
    CIMeC and DISI \\
    University of Trento \\
    Trento, Italy \\
    \texttt{stefano.teso@unitn.it} \\
}

\begin{document}
\maketitle

\begin{abstract}
    There is growing interest in concept-based models (CBMs) that combine high-performance and interpretability by acquiring and reasoning with a vocabulary of high-level concepts.
    A key requirement is that the concepts be interpretable.
    Existing CBMs tackle this desideratum using a variety of heuristics based on unclear notions of interpretability, and fail to acquire concepts with the intended semantics.
    We address this by providing a clear definition of interpretability in terms of alignment between the model's representation and an underlying data generation process,
    and introduce \methods, a new CBM that exploits techniques from disentangled representation learning and open-set recognition to achieve alignment, thus improving the interpretability of the learned concepts.
    We show that \methods, paired with concept-level supervision, achieve better alignment than state-of-the-art approaches while preventing spurious concepts from unintentionally affecting its predictions. The code is available at \href{https://github.com/ema-marconato/glancenet}{https://github.com/ema-marconato/glancenet}.
    \end{abstract}

\section{Introduction}

Concept-based models (CBMs) are an increasingly popular family of classifiers that combine the transparency of white-box models with the flexibility and accuracy of regular neural nets~\citep{alvarez2018towards,li2018deep,chen2019looks,losch2019interpretability,chen2020concept}.
At their core, all CBMs acquire a vocabulary of concepts capturing high-level, task-relevant properties of the data, and use it to compute predictions and produce faithful explanations of their decisions~\citep{rudin2019stop}.

The central issue in CBMs is how to ensure that the concepts are \textit{semantically meaningful} and \textit{interpretable} for (sufficiently expert and motivated) human stakeholders.
Current approaches struggle with this.
One reason is that the notion of interpretability is notoriously challenging to pin down, and therefore existing CBMs rely on different heuristics---such as encouraging the concepts to be sparse~\citep{alvarez2018towards}, orthonormal to each other~\citep{chen2020concept}, or match the contents of concrete examples~\citep{chen2019looks}---with unclear properties and incompatible goals.
A second, equally important issue is \textit{concept leakage}, whereby the learned concepts end up encoding spurious information about unrelated aspects of the data, making it hard to assign them clear semantics~\citep{mahinpei2021promises}.
Notably, even concept-level supervision is insufficient to prevent leakage~\citep{margeloiu2021concept}.

Prompted by these observations, we define interpretability in terms of \textit{alignment}:  learned concepts are interpretable if they can be mapped to a (partially) interpretable data generation process using a transformation that preserves semantics.
This is sufficient to
unveil limitations in existing strategies,
build an explicit link between interpretability and disentangled representations,
and provide a clear and actionable perspective on concept leakage.
Building on our analysis, we also introduce \methods (\acronym), a novel class of CBMs that combine techniques from \textit{disentangled representation learning}~\citep{scholkopf2021toward} and \textit{open-set recognition} (OSR)~\citep{scheirer2012toward} to actively pursue alignment -- and guarantee it under suitable assumptions -- and avoid concept leakage.

\noindent
\textbf{Contributions:}  Summarizing, we:
(\textit{i}) Provide a definition of interpretability as alignment that facilitates tapping into ideas from disentangled representation learning;
(\textit{ii}) Show that concept leakage can be viewed from the perspective of out-of-distribution generalization;
(\textit{iii}) Introduce \methods, a novel class of CBMs that acquire interpretable representations and are robust to concept leakage;
(\textit{iv}) Present an extensive empirical evaluation showing that \methods are as accurate as state-of-the-art CBMs while attaining better interpretability and avoiding leakage.

\begin{figure}[!tb]
    \centering
    \includegraphics[width=0.9\linewidth]{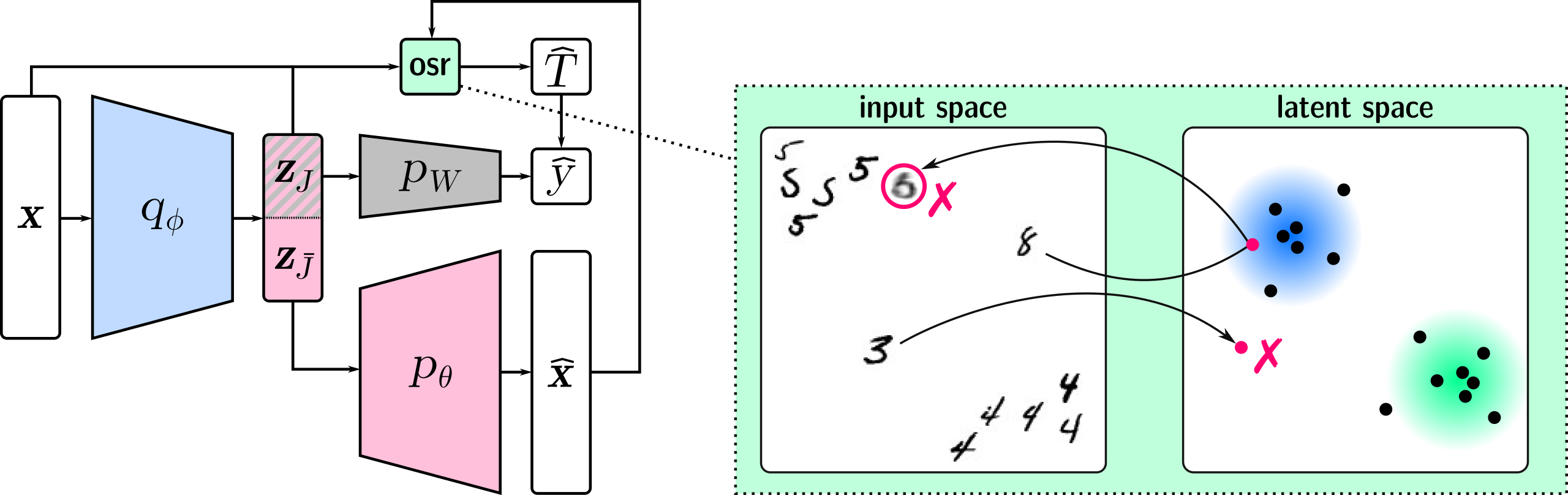}
    \caption{
    \textbf{Left}: Architecture of \methods showing the encoder $q_\phi$, decoder $p_\theta$, classifier $p_W$, and open-set recognition step.
    \textbf{Right}: At test time, \methods prevent leakage by identifying and rejecting out-of-distribution inputs using a combined strategy, shown here for a model trained on digits ``$4$'' and ``$5$'' only:
    the ``$3$'' is rejected as its embedding falls far away from prototypes of the two training classes (colored blobs),
    while the ``$8$'' is rejected as its reconstruction loss is too large.
    }
    \label{fig:architecture}
\end{figure}

\section{Concept-based Models: Interpretability and Concept Leakage}
\label{sec:cbms}

Concept-based models (CBMs) comprise two key elements:
(i) A learned vocabulary of $k$ high-level concepts meant to enable communication with human stakeholders~\citep{kambhampati2021symbols}, and
(ii) a simulatable~\citep{lipton2018mythos} classifier whose predictions depend solely on those concepts.
Formally, a CBM $f: \bbR^d \to [c]$, with $[c] \defeq \{1, \ldots, c\}$, maps instances $\vx$ to labels $y$ by
measuring how much each concept activates on the input, obtaining an activation vector {$\vz(\vx) \defeq ( z_1(\vx), \ldots, z_k(\vx) )^T \in \bbR^k$},
aggregating the activations into per-class scores $s_y(\vx)$ using a linear map~\citep{alvarez2018towards,chen2019looks,chen2020concept}, and then passing these through a softmax, \ie%
\[
    \textstyle
    s_y(\vx) \defeq \sum_j w_{yj} \cdot z_j(\vx),
    \qquad 
    p(y \ |\ \vx) \defeq \softmax( \vs(\vx) )_y.
\]
Each weight $w_{yj} \in \bbR$ encodes the relevance of concept $z_j$ for class $y$.
The activations themselves are computed in a black-box manner, often leveraging pre-trained embedding layers, but learned so as to capture interpretable aspects of the data using a variety of heuristics, discussed below.

Now, \textit{as long as the concepts are interpretable}, it is straightforward to extract human understandable local explanations disclosing how different concepts contributed to any given decision $(\vx, y)$ by looking at the concept activations and their associated weights, thus abstracting away the underlying computations.
This yields explanations of the form
$
    \{(w_{yj}, \, z_j(\vx)) : j \in [k]\}
    \label{eq:cbm-explanations}
$
that can be readily summarized\footnote{For instance, by pruning those concepts that have little effect on the outcome to simplify the presentation.} and visualized~\citep{hase2020evaluating,guidotti2018survey}.
Importantly, the score of class $y$ is conditionally independent from the input $\vx$ given the corresponding explanation,
\ie $s_y(\vx) \indep \vx \mid \expl(\vx, y)$,
ensuring that the latter is faithful to the model scores.  \methods inherit all of these features.

\noindent
\textbf{Heuristics for interpretability.}  Crucially, CBMs are only interpretable insofar as their concepts are.  Existing approaches implement special mechanisms to this effect, often pairing a traditional classification loss (such as the cross-entropy loss) with an auxiliary regularization term.

\citet{alvarez2018towards} acquire concepts using an autoencoder augmented with a sparsification penalty encouraging distinct concepts to activate on different instances.
\citet{chen2020concept} apply geometric transforms to learn mutually orthonormal concepts that thus encode complementary information and attain comparable activation ranges.
These mechanisms -- sparsity and orthogonality, respectively -- alone cannot prevent capturing features that are not semantic in nature.

A second group of CBMs tackle this issue by constraining the concepts to match \textit{concrete} cases, in the hope that these are better aligned with human intuition~\citep{kim2016examples}.
For instance, prototype classification networks~\citep{li2018deep}, part-prototype networks~\citep{chen2019looks}, and related approaches~\citep{rymarczyk2020protopshare,nauta2021neural,singh2021these} model concepts using prototypes in embedding space that perfectly match training examples or parts thereof.
Depending on the embedding space, which ultimately determines the distance to the prototypes, concepts learned this way may activate on elements unrelated to the example they match, leading to unclear semantics~\citep{hoffmann2021looks}.

Closest to our work, concept bottleneck models (CBNMs)~\citep{koh2020concept,losch2019interpretability} align the concepts using concept-level supervision -- possibly obtained from a separate source, like ImageNet~\citep{deng2009imagenet} -- either sequentially or in tandem with the top-level dense layer.
From a statistical perspective, this seems perfectly sensible:  if the supervision is unbiased and comes in sufficient quantity, and the model has enough capacity, this strategy \textit{appears} to guarantee the learned and ground-truth concepts to match.

\noindent
\textbf{Concept leakage in concept-bottleneck models.}  Unfortunately, concept-level supervision is \textit{not} sufficient to guarantee interpretability.
\citet{mahinpei2021promises} have demonstrated that concepts acquired by CBNMs pick up spurious properties of the data.
In their experiment, they learn two concepts $z_4$ and $z_5$, meant to represent the $4$ and $5$ MNIST digits, using concept-level supervision, and then show that -- surprisingly -- these concepts can be used to classify all \textit{other} digits (\ie MNIST images that are neither $4$'s nor $5$'s) as even or odd significantly better than random guessing.  This phenomenon, {whereby learned concepts unintentionally capture information about unobserved concepts,}
is known as \textit{concept leakage}.

Intuitively, leakage occurs because  in CBNMs the concepts end up unintentionally capturing distributional information about unobserved aspects of the input, failing to provide well-defined semantics.
However, a clear definition of leakage is missing, and so are strategies to prevent it.  In fact, separating concept learning from classification and increasing the amount of supervision for the observed concepts (here, $4$ and $5$) is not enough~\citep{margeloiu2021concept}.
A key contribution of our paper is showing that leakage can be understood from the perspective of domain shift and dealt with using open-set recognition~\citep{scheirer2012toward}.

\section{Disentangling Interpretability and Concept Leakage}
\label{sec:alignment-and-leakage}

The main issue with heuristics used by CBMs is that they are based on unclear notions of interpretability.
In order to develop effective algorithms, we propose to view interpretability as a form of \textit{alignment} between the machine's representation and that of its user.
This enables us to
identify conditions under which interpretability can be achieved,
build links to well-understood properties of representations, and leverage state-of-the-art learning strategies.

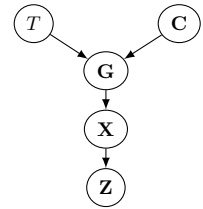
\begin{wrapfigure}{r}{0.275\linewidth}
    \centering
    \vspace{-1em}
    \begin{minipage}[c]{0.5\linewidth}
        \centering
        \begin{tikzpicture}[
            scale=0.75,
            transform shape,
            node distance=.35cm and .75cm,
            minimum width={width("Xtil")+2pt},
            minimum height={width("Xtil")+2pt},
            mynode/.style={draw,ellipse,align=center}
        ]
            \node[mynode] (G) {$\vG$};
            \node[mynode, above right=of G] (C) {$\vC$};
            \node[mynode, above left=of G] (T) {$T$};
            \node[mynode, below=of G] (X) {$\vX$};
            \node[mynode, below=of X] (Z) {$\vZ$};

            \path (C) edge[-latex] (G)
            (G) edge[-latex] (X)
            (X) edge[-latex] (Z)
            (T) edge[-latex] (G);

        \end{tikzpicture}
    \end{minipage}
    \caption{The data generation process.}
    \label{fig:generative-process}
\end{wrapfigure}
\noindent
\textbf{Interpretability.}  We henceforth focus on the (rather general) generative process shown in~\cref{fig:generative-process}:
the observations $\vX \in \bbR^d$ are caused by $n$ generative factors $\vG \in \bbR^n$, themselves caused by a set of confounds $\vC$ (including the label $Y$~\citep{scholkopf2012causal}).  Notice that the generative factors \textit{can} be statistically dependent due to the confounds $\vC$, but as noted by \citet{suter2019robustly}, the total causal effect \cite[Def.~6.12]{peters2017elements} between $G_i$ and $G_j$ is zero for all $i \ne j$.
The generative factors capture all information necessary to determine the observation~\citep{suter2019robustly,reddy2021causally}, so the goal is to learn concepts $\vZ \in \bbR^k$ that recover them.
The variable $T$ is also a confounding factor, but it is kept separate from $\vC$ as it relates to concept leakage, and will be formally introduced later on.

We posit that a (learned) representation is only interpretable if it supports \textit{symbolic communication} between the model and the user, in the sense that it shares the same (or similar enough) semantics to the user's representation.  The latter is however generally unobserved.
We therefore make a second, critical assumption that \textit{some} of the generative factors $\vG_I \subseteq \vG$ are interpretable to the user, meaning that they can be used as a proxy for the user's internal representation.
Naturally, not all generative factors are interpretable~\citep{gabbay2021image}, but in many applications some of them are.  For instance, in dSprites~\citep{dsprites17} the generative factors encode the position, shape and color of a 2D object, and in CelebA~\citep{liu2015faceattributes} the hair color and nose size of a celebrity.  Human observers have a good grasp of such concepts.

\noindent
\textbf{Interpretability as alignment.}  Under this assumption, if the variables $\vZ_J \subseteq \vZ$ are \textit{aligned} to the generative factors $\vG_I$ by a map $\alpha: \vg \mapsto \vz_J$ that preserves semantics, they are themselves interpretable.
Now, defining what a semantics-preserving map should look like is challenging, but constructing one is not:  the identity is clearly one such map, and so are maps that permute the indices and independently rescale the individual variables.
One desirable property is that $\alpha$ does not ``mix'' multiple $G$'s into a single $Z$.  E.g., if $Z$ blends together head tilt, hair color, and nose size, users will have trouble pinning down what it means.\footnote{The converse is not true:  interpretable concepts with \textit{compatible} semantics can be mixed without compromising interpretability.  E.g., rotating a coordinate system gives another intuitive coordinate system.  Our point is that conservatively avoiding mixing helps to preserve semantics.}
This property can be formalized in terms of \textit{disentanglement}~\citep{eastwood2018framework,suter2019robustly,scholkopf2021toward}.
This is however insufficient:  we wish the map between $G_i$ and its associated factor $Z_j$ to be ``simple'', so as to \textit{conservatively} guarantee that it preserves semantics.  This makes alignment strictly stronger than disentanglement.

Motivated by these desiderata, we say that $\vZ_J$ is \textit{aligned} to $\vG_I$ { if it satisfies}:
\begin{itemize}[leftmargin=1.25em]

    \item[(\textit{i})] {\textbf{Disentanglement.}} There exists an injective map between indices $\pi: [n_I] \to [k]$, where $[n_I]$ identifies the subset of generative factors indexes in $\vG_I$,  such that, for all $i, i' \in [n_I]$, $i \ne i'$, and $j = \pi(i)$, it holds that fixing $G_i$ is enough to fix $Z_j$ regardless of the value taken by the other generative factors $G_{i'}$, and

    \item[(\textit{ii})] {\textbf{Monotonicity.}} The map $\alpha$ can be written as {$\alpha(\vg) = (\mu_1(g_{\pi(1)}), \ldots, \mu_n(g_{\pi(n_I)}))^T$,} where the $\mu_i$'s are monotonic transformations.  This holds, for instance, for linear transformations of the form {$A \ (g_{\pi(1)}, \ldots, g_{\pi(n_I)})^T$ }, where $A \in \bbR^{n_I \times k}$ is a matrix with no non-zero off-diagonal entries.  This second requirement can be relaxed depending on the application.

\end{itemize}
Notice that we do not require each $G_i$ to map to a \textit{single} $Z_j$ (a property known as \textit{completeness}~\citep{eastwood2018framework}):  $\vZ_J$ is interpretable even if it contains multiple -- perhaps slightly different, but aligned -- transformations of the same $G_i$.

\noindent
\textbf{Measuring alignment with DCI.}  Disentanglement can be measured in a number of ways~\citep{zaidi2020measuring}, but most of them provide little information about how simple the map $\alpha$ is.  In order to estimate alignment, we repurpose DCI, a measure of disentanglement introduced by~\citet{eastwood2018framework}, see also \cref{sec:dci-framework}.  According to this metric, a representation $\vZ_J$ is disentangled if there exists a regressor that, given $\vz_J$, can predict $\vg_I$ with high accuracy using few $z_i$'s to predict each $g_i$.  Following~\citep{eastwood2018framework}, we use a linear regressor with parameters $B \in \bbR^{k \times n_I}$ on the test set -- assuming that it is annotated with the interpretable generative factors and corresponding learned representations -- and then measure how diffuse the weights associated to each latent factor are.  We do this by normalizing them and computing their average Shannon entropy over all $G_i$'s, \ie%
\[
    \textstyle
    -  \sum_{j \in [k]} \rho_j \left(
        \sum_{i \in [n_I]}
            \bar{b}_{ji} \log \bar{b}_{ji}
    \right),
    \quad
    \text{where $\bar{b}_{ji} = b_{ji} / \sum_{j' \in [k]} b_{j'i}$}
    \quad \text{and} \quad
    \rho_j = \sum_i b_{ji} / \sum_{j' i} b_{j' i}
\]
Hence, DCI gauges the degree of mixing that a linear map can attain using the learned representation $\vZ$, and as such it indirectly measures alignment, with $B$ approximating the inverse of $A$.

\noindent
\textbf{Achieving alignment with concept-level supervision.}  It has been shown that disentanglement cannot be achieved in the purely unsupervised setting~\citep{locatello2019challenging}.  This immediately entails that alignment is also impossible in that setting, highlighting a core limitation of approaches like self-explainable neural networks~\citep{alvarez2018towards}.
However, disentanglement can be attained if supervision about the generative factors is available, even only for a small percentage of the examples~\citep{locatello2020disentangling}.
As a matter of fact, supervision is used in representation learning to achieve \textit{identifiability}, a stronger condition than -- and that entails both of -- disentanglement \textit{and} alignment~\citep{khemakhem2020variational}.
Thus, following CBNMs, we seek alignment by leveraging concept-level supervision.

\noindent
\textbf{Interpretability and concept leakage.}  Intuitively, concept leakage occurs when a model is trained on a data set on which:
\begin{itemize}[leftmargin=1.25em]

    \item[(\textit{i})] Some generative factors $\vG_V \subset \vG$ \textbf{vary}, while the others $\vG_F = \vG \setminus \vG_V$ are \textbf{fixed}, and

    \item[(\textit{ii})]  The two groups of factors are \textbf{statistically dependent}.

\end{itemize}
For instance, in the even vs.\@ odd experiment {$4$ and $5$ play the role of $\vG_{V}$ and the other digits of $\vG_F$}.
CBNMs with access to supervision on $\vG_V$ tend to acquire a latent representation that approximates these factors. {But, because of (\textit{ii}), this representation correlates with the fixed factors $\vG_F$.}
This immediately explains why additional supervision on $\vG_V$ cannot prevent leakage, but rather has the opposite effect: {the better a latent representation matches $\vG_V$, the more information it conveys about $\vG_F$.}

In contrast with previous assessments~\citep{mahinpei2021promises,margeloiu2021concept}, we observe that {this phenomenon} can be viewed as a special form of domain shift:  the training examples are sampled from a ground-truth distribution $p(\vX, \vG \mid T=1)$ in which $\vG_F$ is approximately fixed, \eg $p(\vG_F \mid T=1) = \delta(\vg_F')$ for some vector $\vg_F'$, while in the test set, the data is sampled from a different distribution $p(\vX, \vG \mid T=0)$ in which $\vG_F$ is no longer fixed. {In the MNIST task, for instance, when $T=1$ no concept besides $4$ and $5$ can occur, while all concepts \textit{except} $4$ and $5$ can occur when $T=0$.}  Here, $T \in \{0, 1\}$ selects between training and test distribution, see~\cref{fig:generative-process}. 
{Now, CBMs have no strategy to cope with domain shift and thus cannot disambiguate between known training and unknown test concepts.}

Motivated by this, we propose then to tackle concept leakage by designing a CBM specifically equipped with strategies for detecting {-- at \textit{inference} time --} instances that do not belong to the training distribution using OSR~\citep{scheirer2012toward}.  The idea is to estimate the value of the variable $T$ at inference time, essentially predicting whether an input was sampled from a distribution similar enough to the training distribution, and therefore can be handled by a model learned on this distribution, or not.  This strategy proves very effective in practice, as shown by our empirical evaluation (\cref{sec:q3-leakage}).

\section{Addressing Alignment and Leakage with \methods}
\label{sec:methods}

\methods combine a VAE-like architecture~\citep{kingma2014auto,rezende2014stochastic} for learning disentangled concepts with a prior and classifier designed for open-set prediction~\citep{sun2020conditional}.
In order to accommodate for non-interpretable factors, the latent representation of \methods $\vZ$ is split into two:
(i) $k$ concepts $\vZ_J$, aligned to the \textit{interpretable} generative factors $\vG_I$, that are used for prediction, and
(ii) $\bar{k}$ \textit{opaque} factors $\vZ_{\bar{J}}$ that are only used for reconstruction.
Specifically, a \method comprises an encoder $q_\phi(\vZ\mid\vX)$ and a decoder $p_\theta(\vX\mid\vZ)$, both parameterized by deep neural networks, as well as a classifier $p_W(Y \mid \vZ_J)$ feeding off the interpretable concepts only.  The overall architecture is shown in~\cref{fig:architecture}.

Following other CBMs, the classifier is implemented using a dense layer with parameters $W \in \bbR^{v \times k}$ followed by a softmax activation, \ie $p_W(Y \mid \vz_J) \defeq \softmax(W \vz_J)$, and the most likely label is used for prediction.
The class distribution is obtained by marginalizing over the encoder's distribution:%
\[
    p(Y\mid\vx) \defeq \bbE_{q_\phi(\vz\mid\vx)} [ p(Y\mid\vz,\vx) ] = \bbE_{q_\phi(\vz\mid\vx)} [ p_W(Y\mid\vz_J) ]
    \label{eq:predict-by-marginalizing}
\]
Equality holds because $Y \indep \vX \mid \vZ_J$.  In order to expedite the computation, we follow the general practice of approximating the integral as
$
    \softmax( W \, \bbE_{q_\phi(\vz\mid\vx)}[\vz_J] ) = \softmax( W [ \vmu_\phi(\vx) ]_J )
$.

In contrast to regular VAEs, \methods associate each class to a prototype in latent space through the prior $p(\vZ\mid\vY)$, which is conditioned on the class and modelled as a \textit{mixture of gaussians} with one component per class.
The encoder, decoder, and prior are fit on data so as to maximize the evidence lower bound (ELBO)~\citep{kingma2019introduction}, defined as $\bbE_{p_{\dataset}(\vx, y)} [\calL(\theta, \vx, y; \beta)]$ with:
\[
    \calL(\theta, \vx, y; \beta) \defeq
        \bbE_{q_\phi(\vz\mid\vx)}[ \log p_\theta(\vx\mid\vz) + \log p_W(y\mid\vz_J) ] - \beta \cdot \KL(q_\phi(\vz\mid\vx)\,\|\,p(\vz\, | \, y))
    \label{eq:elbo}
\]
Here, $p_{\dataset}(\vx, y)$ is the empirical distribution of the training set $\dataset = \{ (\vx_i, y_i) : i = 1, \ldots, m \}$.
The first term of~\cref{eq:elbo} is the likelihood of an example after passing it through the encoder distribution.

The second term penalizes the latent vectors based on how much their distribution differs from the prior and encourages disentanglement.
As mentioned in~\cref{sec:alignment-and-leakage}, learning disentangled representations is impossible in the unsupervised i.i.d. setting~\citep{locatello2019challenging}. 
Following~\citet{locatello2020disentangling}, and similarly to CBNMs, we assume access to a (possibly separate) data set $\widetilde{\dataset} = \{ (\vx_\ell, \vg_{I,\ell}) \}$ containing supervision about the \textit{interpretable} generative factors $\vG_I$ and integrate it into the ELBO by replacing the per-example loss $\calL$ in~\cref{eq:elbo} with:
\begin{align}
    {\bbE_{p_\mathcal{D}(\vx, y) } }\big[
    \calL(\theta, \vx, y; \beta) \big] + \gamma \cdot  \bbE_{p_{\Tilde{\dataset}}(\vx, \vg) } \bbE_{q_\phi(\vz\mid\vx)}  \big[  \Omega ( \vz, \vg)   \big]
    \label{eq:elbo+int}
\end{align}
where $\gamma > 0$ controls the strength of the concept-level supervision.
Following \cite{locatello2020disentangling}, the term $\Omega (\vz, \vg) $ penalizes encodings sampled from $q_\phi(\vz \, | \, \vx)$ for differing from the annotation $\vg$.
Specifically, we implement this term using the average cross-entropy loss $\Omega ( \vz , \vg) \defeq -\sum_k g_k \log \sigma(z_k) + (1-g_k) \log (1- \sigma(z_k))$, where the annotations $g_k$ are rescaled to lie in $[0,1]$ and $\sigma$ is the sigmoid function.

\noindent
\textbf{Dealing with concept leakage.}  In order to tackle concept leakage, \methods integrate the OSR strategy of~\citet{sun2020conditional}, indicated in \cref{fig:architecture} by the ``osr'' block.
This strategy identifies out-of-class inputs by considering the class prototype $\mu_y := \bbE_{p(\vz | y)}[\vz] $ in $\bbR^k$ defined by the prior distribution and the decoder $p_\theta(\vx | \vz)$.  Recall that the prior is fit jointly with the encoder, decoder, and classifier by optimizing the ELBO.  Once learned, a \method uses the training data to estimate:
(\textit{i}) a distance threshold $\eta_y$, which defines a spherical subset in the latent space ${\calB}_y = \{ \vz: \, \norm{ \mu_y -\vz  } < \eta_y$ \} centered around the prototype of class $y$ (\ie the mean of the corresponding Gaussian mixture component), and
(\textit{ii}) a maximum threshold on the reconstruction error $\eta_{thr}$.
If new data points have reconstruction error above $\eta_{thr}$ or they do not belong to any subset ${\calB}_y$, they are inferred as open-set instances, \ie $\hat{T} = 0$.
In practice, we found that choosing the thresholds as to include 95\% of training examples to work well in our experiments. {Further details are available in Appendices \ref{sec: implementation details} and \ref{sec: openset appendix}.}

\subsection{Benefits and Limitations}

\methods can naturally be combined with different VAE-based architectures for learning disentangled representations~\cite{esmaeili2019structured}, including $\beta$-TCVAEs~\citep{chen2018isolating}, InfoVAEs~\citep{Zhao_Song_Ermon_2019}, DIP-VAEs~\citep{kumar2018variational}, and JL1-VAEs~\citep{rhodes2021local}.  Since our experiments already show substantial benefits for \methods building on $\beta$-VAEs~\citep{higgins2016beta}, we leave a detailed study of these extensions to future work.

Like CBNMs, \methods foster alignment by leveraging supervision on the interpretable generative factors~\citep{locatello2020disentangling}, possibly derived from an external data set~\citep{koh2020concept}.
However, \methods can be readily adapted to a variety of different kinds of supervision used for VAE-based models, including \textit{partially} annotated examples~\citep{gabbay2021image}, group information~\citep{bouchacourt2018multi}, pairings~\citep{shu2020weakly,locatello2020weakly} and other kinds of weak supervision~\citep{gabbay2019latent,chen2020weakly}, as well as feedback from a domain expert~\citep{stammer2021interactive}.
On the other hand, CBNMs are incompatible with these approaches.

One limitation inherited from VAEs by \methods is the assumption that the interpretable generative factors are disentangled from each other~\citep{suter2019robustly}.  In practice, \methods work even when this does not hold (as in our even vs.\@ odd experiment, see~\cref{sec:q3-leakage}).  However, one direction of future work is to integrate ideas from hierarchical disentanglement~\citep{ross2021benchmarks}.

\section{Empirical Evaluation}
\label{sec: experiments}

In this section, we present results on several tasks showing that \methods outperform CBNMs~\citep{koh2020concept} in terms of alignment and robustness to leakage, while achieving comparable prediction accuracy.
All experiments were implemented using Python 3 and Pytorch~\citep{paszke2019pytorch} and run on a server with 128 CPUs, 1TiB RAM, and 8 A100 GPUs.
\methods were implemented on top of the {\tt disentanglement-pytorch}~\citep{abdi2019variational} library.
All alignment and disentanglement metrics were computed with {\tt disentanglement\_lib}~\citep{locatello2019challenging}.
Code for the complete experimental setup is available on GitHub at the link: \href{https://github.com/ema-marconato/glancenet}{https://github.com/ema-marconato/glancenet}.
Additional details on architectures and hyperparameters can also be found in the Supplementary Material.

\begin{figure}[!t]
    \centering
    {\footnotesize\sc
    \begin{tabular}{cccc}
        & Accuracy & Alignment & Explicitness \\    
        \rotatebox{90}{\hspace{2em} dSprites} &
        \includegraphics[width=0.275\linewidth]{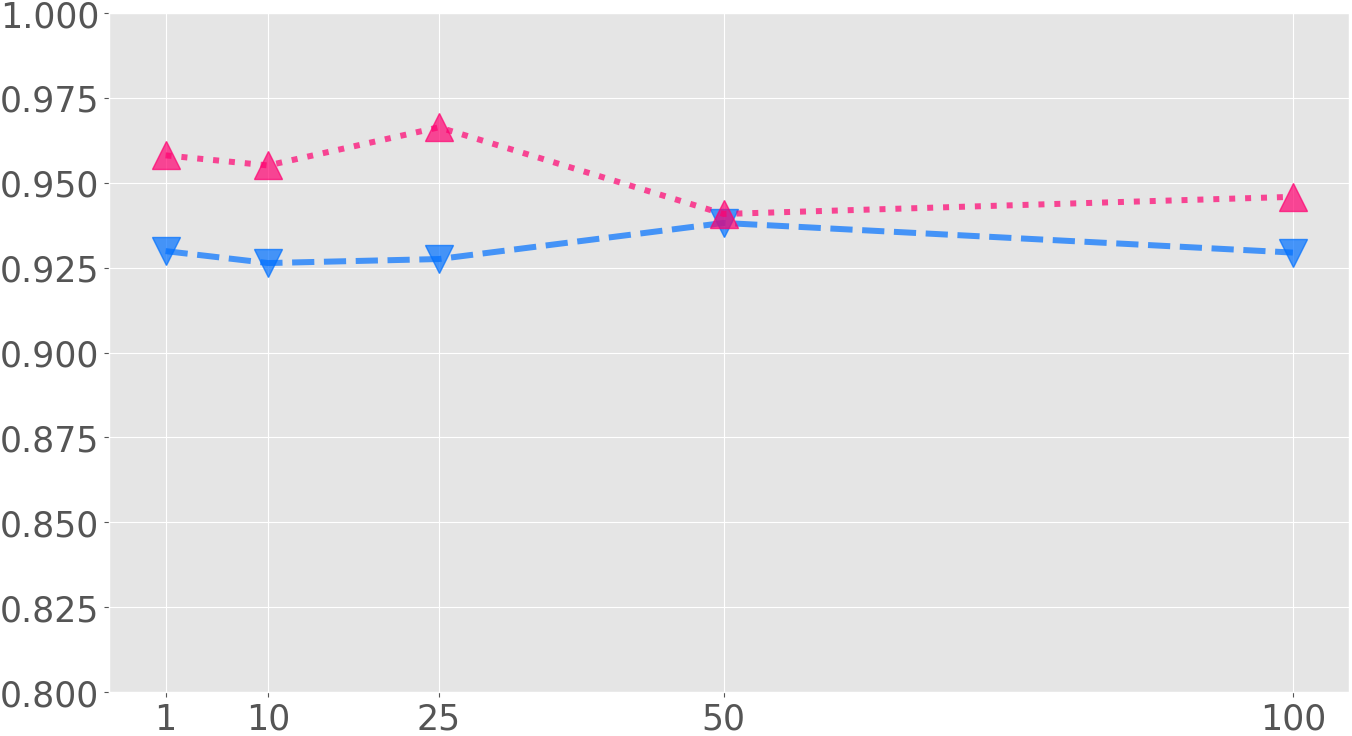} &
        \includegraphics[width=0.275\linewidth]{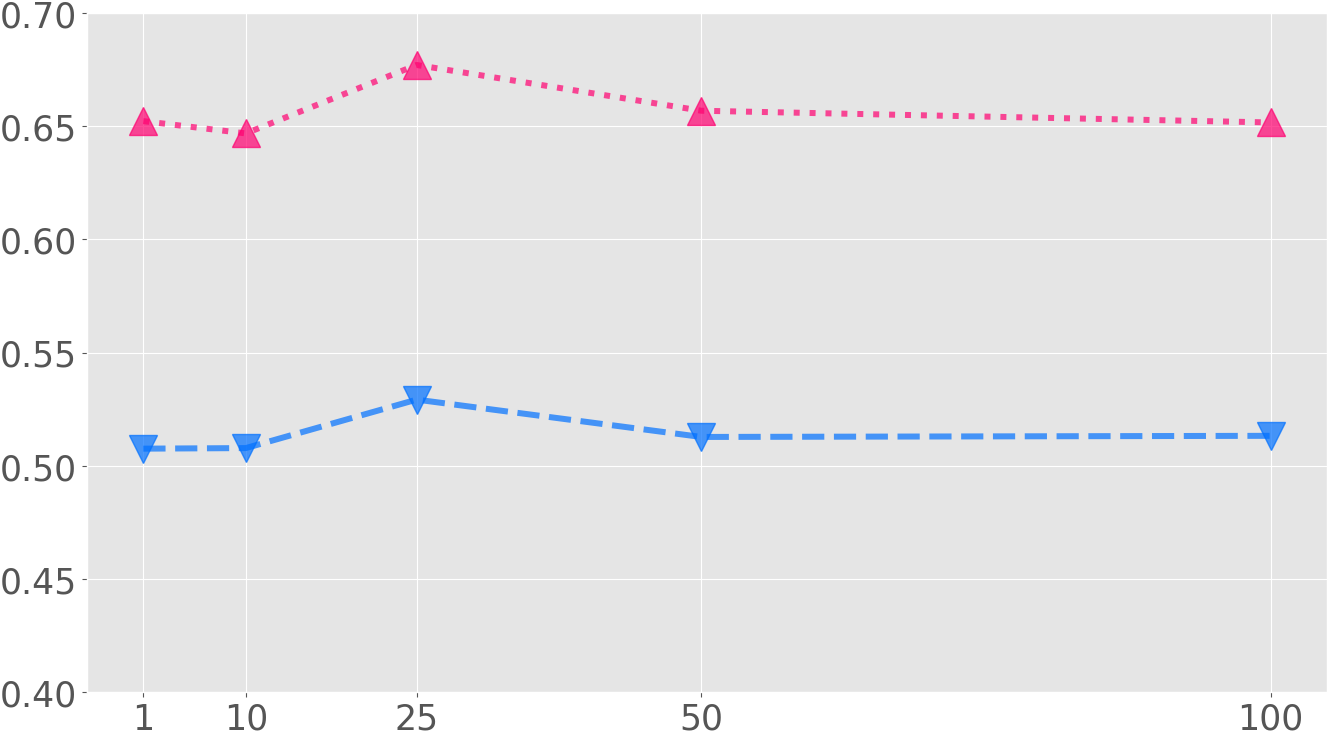} &
        \includegraphics[width=0.275\linewidth]{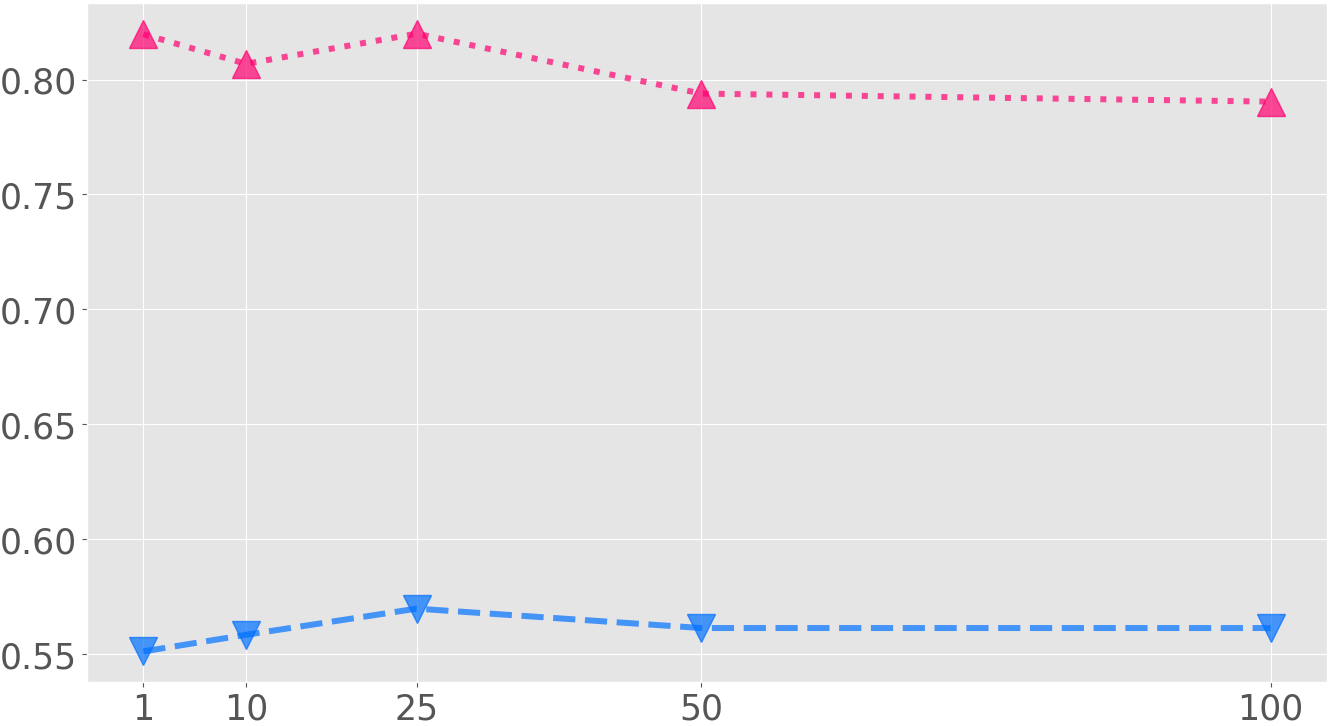}
        \\
        \rotatebox{90}{\hspace{2em} MPI3D} &
        \includegraphics[width=0.275\linewidth]{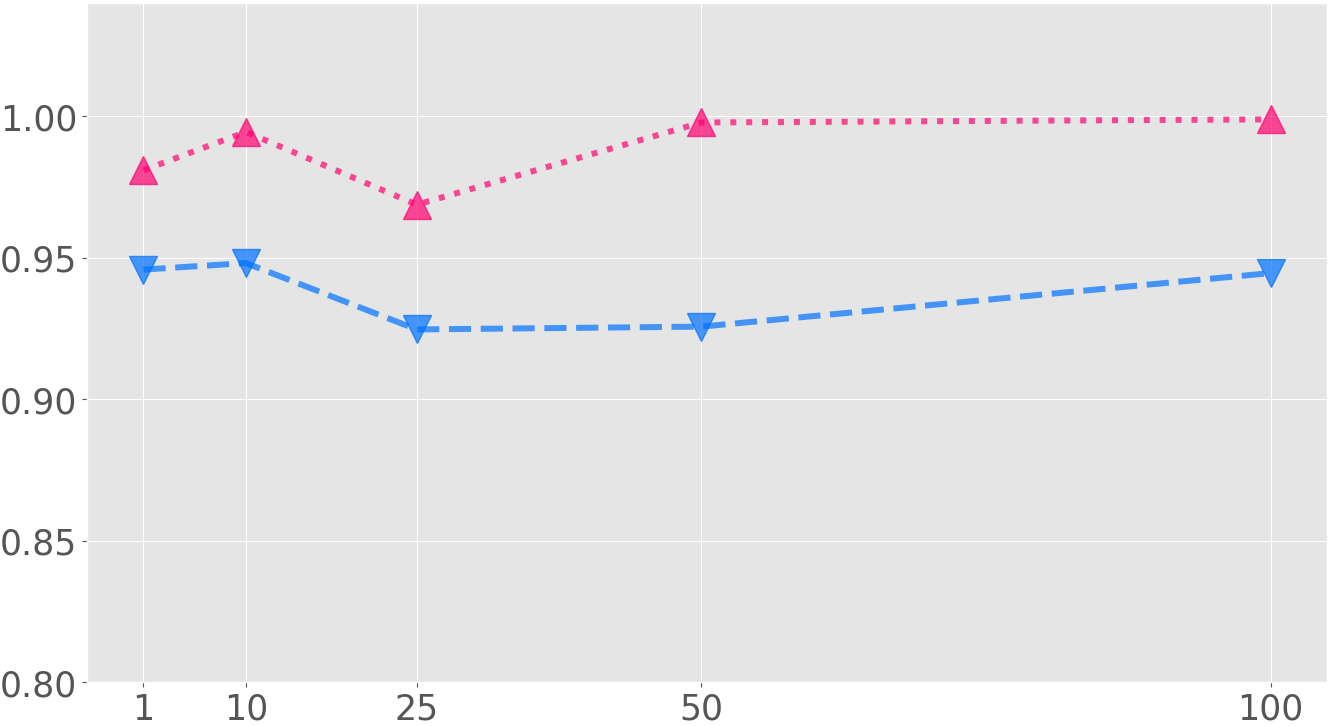} &
        \includegraphics[width=0.275\linewidth]{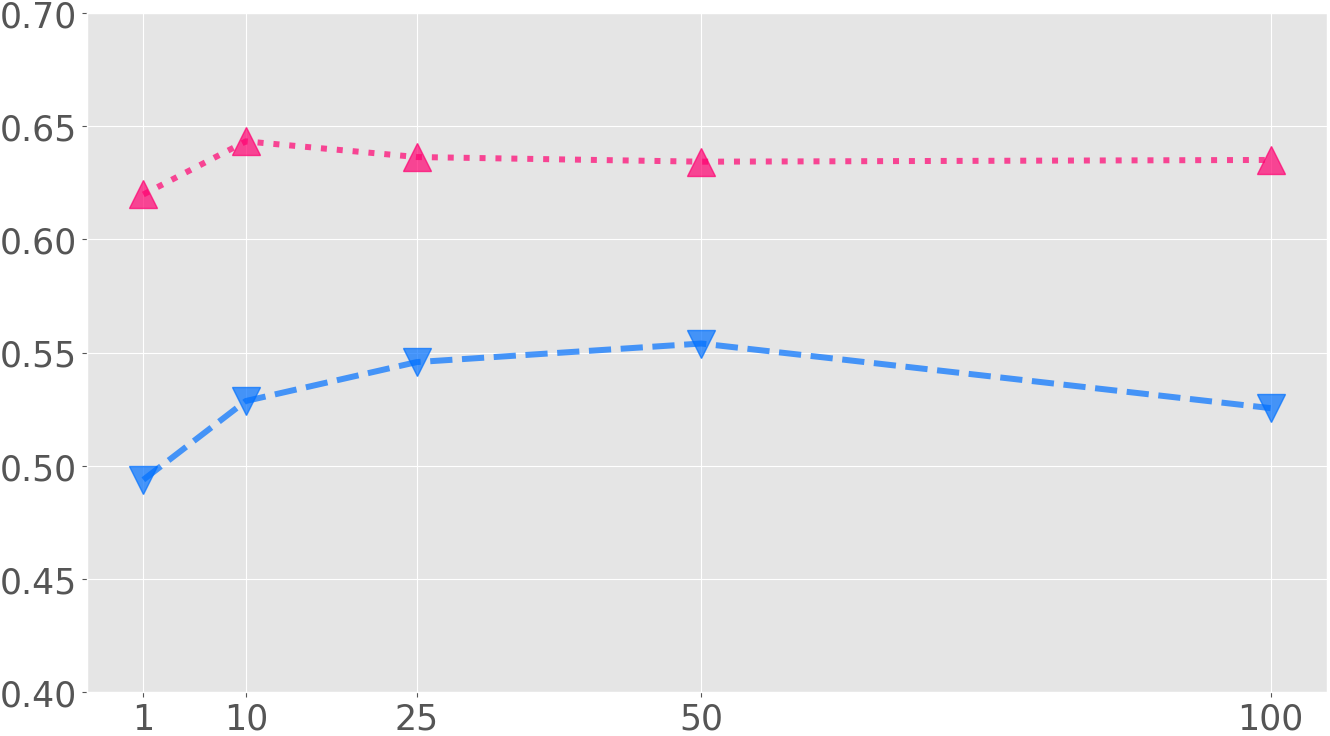} &
        \includegraphics[width=0.275\linewidth]{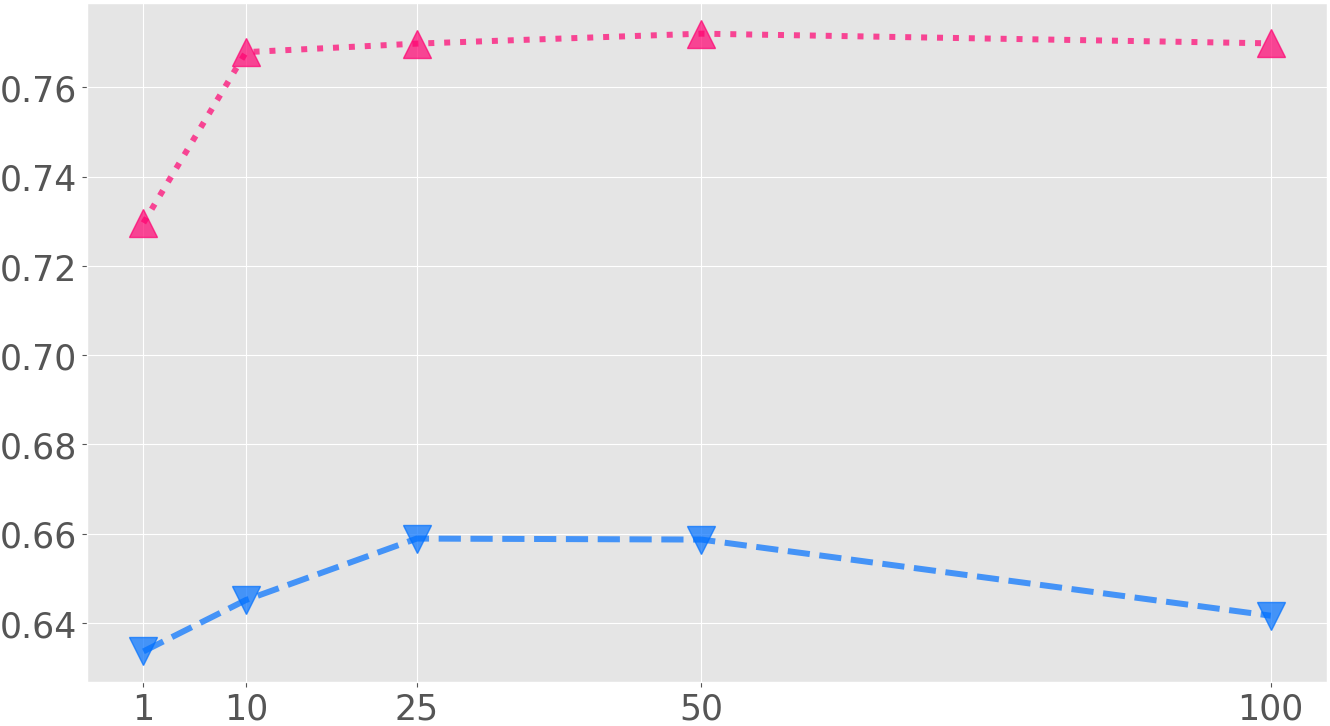}
        \\
        \rotatebox{90}{\hspace{2em} CelebA} &
        \includegraphics[width=0.275\linewidth]{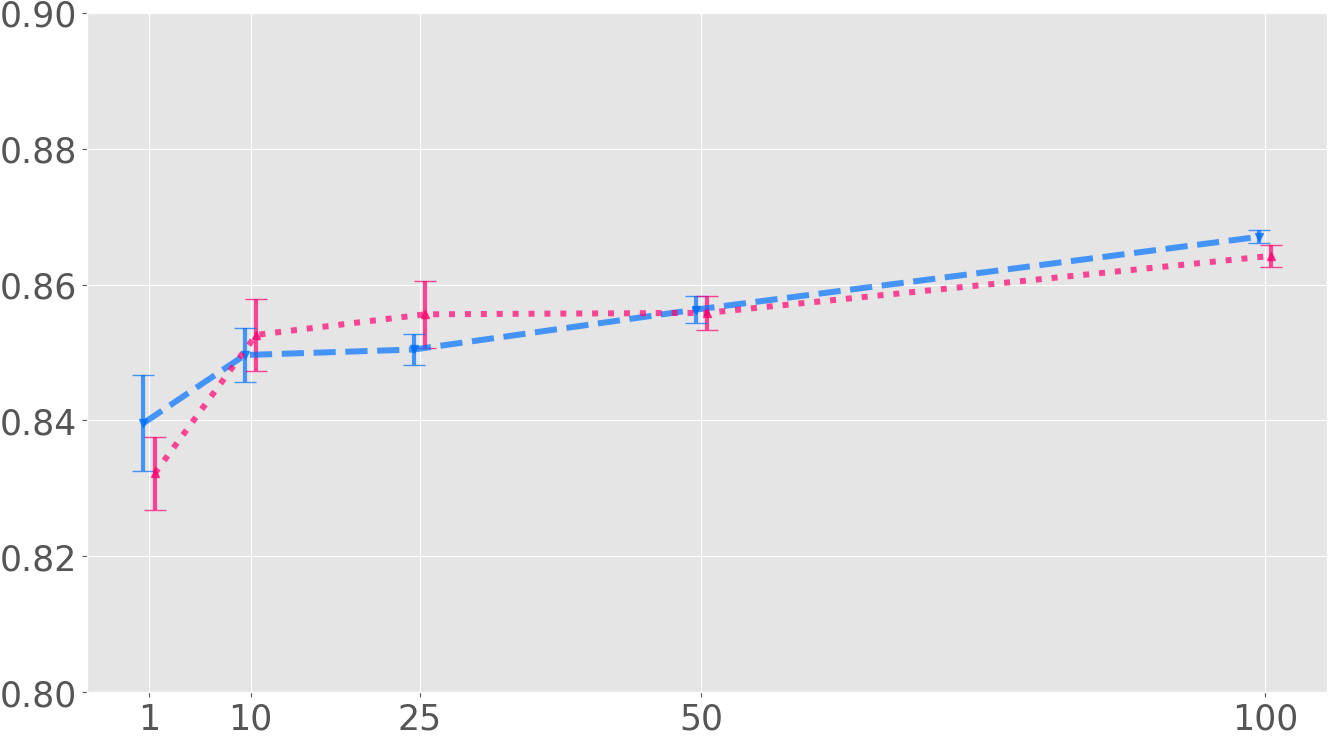} &
        \includegraphics[width=0.275\linewidth]{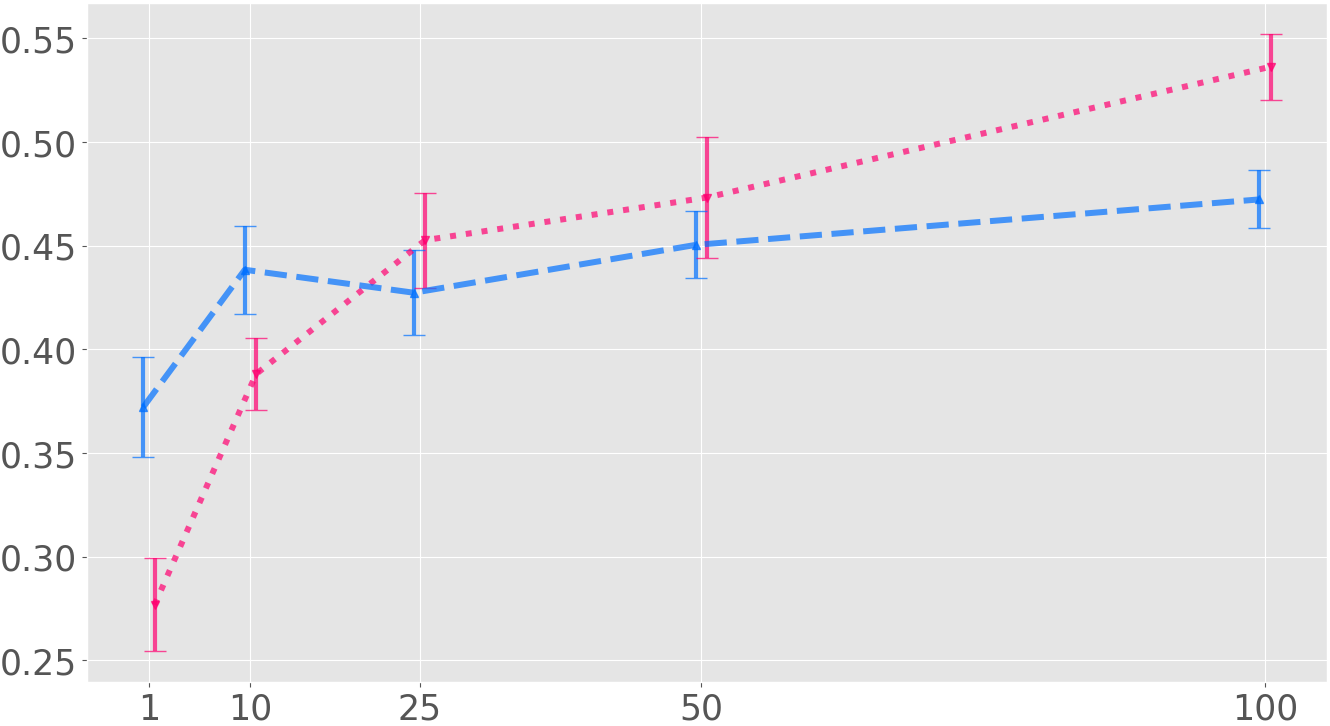} &
        \includegraphics[width=0.275\linewidth]{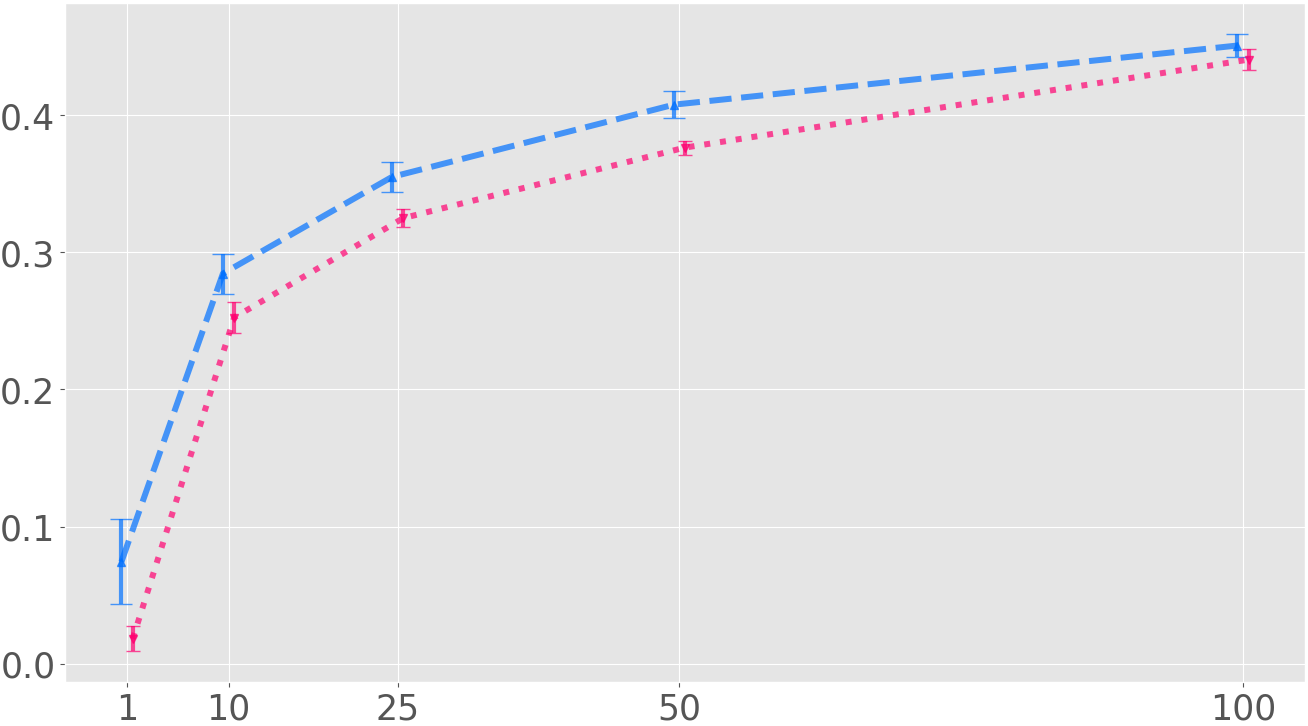}
    \end{tabular}
    }
    \caption{\textbf{\methods are better aligned than CBNMs.}
    Each row is a data set and each column reports a different metric.  The horizontal axes indicate the $\%$ of training examples for which supervision on the generative factors is provided.  Remarkably, in all data sets \textbf{\textcolor{glancecolor}{\methods}} achieve substantially better alignment than \textbf{\textcolor{cbnmcolor}{CBNMs}} for the same amount of supervision, and achieve comparable accuracy in $14$ cases out of $15$.}
    \label{fig:q1}
\end{figure}

\subsection{\methods achieve better alignment than CBNMs}
\label{sec:q1}

In a first experiment, we compared \methods with CBNMs on three classification tasks for which supervision on the generative factors is available.  In order to evaluate the impact of this supervision on the different competitors, we varied the amount of training examples annotated with it from $1\%$ to $100\%$.  For each increment, we measured prediction performance using accuracy, and alignment using the linear variant of DCI~\citep{eastwood2018framework} discussed in~\cref{sec:alignment-and-leakage}.

\noindent
\textbf{Data sets.}  We carried out our evaluation on two data sets taken from the disentanglement literature and a very challenging real-world data set.
\noindent
\underline{\textit{dSprites}}~\citep{dsprites17} consists of $64 \times 64$ black-and-white images of sprites on a flat background, where each sprite is determined by one categorical and four generative factors, namely \texttt{shape}, \texttt{size}, \texttt{rotation}, \texttt{position\_x}, and \texttt{position\_y}.  The images were obtained by discretizing and enumerating the generative factors, for a total of $3 \times 6 \times 40 \times 32 \times 32 $ images.
\noindent
\underline{\textit{MPI3D}}~\citep{NEURIPS2019_d97d404b} consists of $64\times 64$ RGB rendered images of 3D shapes held by a robotic arm.  The generative factors are \texttt{object\_color}, \texttt{object\_shape}, \texttt{object\_size}, \texttt{camera\_height}, \texttt{background\_color}, and the \texttt{horizontal} and \texttt{vertical} position of the arm. The data contains $6 \times 6 \times 2 \times 3 \times 4 \times 40 \times 40$ examples.
\noindent
\underline{\textit{CelebA}}~\citep{liu2015faceattributes} is a collection of $178\times 218$ RGB images of over 10k celebrities, converted to $64 \times 64$ by first cropping them to $178\times 178$ and then rescaling. 
Images are annotated with 40 binary generative factors including hair color, presence of sunglasses, \etc  Since we are interested in measuring alignment, we considered only those $10$ factors that CBNMs can fit well (in the Appendix).  We also dropped all those examples for which hair color is not unique (e.g., annotated as both {\tt blonde} and {\tt black}), obtaining approx. $127k$ examples.  CelebA is more challenging than dSprites and MPI3D, as it does not include all possible factor variations and the generative factors -- although disentangled -- are insufficient to completely determine the contents of the images. For dSprites and MPI3D, we used a random 80/10/10 train/validation/test split, while for CelebA we kept the original split~\cite{liu2015faceattributes}.

We generated the ground-truth labels $y$ as follows.
For dSprites, we labeled images according to a random but fixed linear separator defined over the \textit{continuous} generative factors, chosen so as to ensure that the classes are balanced.
For MPI3D and CelebA, we focused on the \textit{categorical} factors instead.  Specifically, we clustered all images using the algorithm of~\citep{Huang97clusteringlarge}, for a total of $10$ and $4$ clusters for MPI3D and CelebA respectively, and then labeled all examples based on their reference cluster.  This led to slightly unbalanced classes containing different percentages of examples, ranging from $5\%$ to $16\%$ in MPI3D and from $21\%$ to $29\%$ in CelebA.

\noindent
\textbf{Architectures.}  For dSprites and MPI3D, we implemented the encoder as a six layer convolutional neural net, while for CelebA we adapted the convolutional architecture of~\citet{ghosh2020deterministic}.  We employed a six layer convolutional architecture for the decoder in all cases, for simplicity, as changing it did not lead to substantial differences in performance.  In all cases, as for all CBMs (see~\cref{sec:cbms}), the classifier was implemented as a dense layer followed by a softmax activation.  The very same architectures were used for both \methods and CBNMs, for fairness. For each data set, we chose the latent space dimension as the total number of generative factors, where categorical ones are one hot encoded. In particular, we used 7 latent factors for dSprites, 21 for MPI3D and 10 for CelebA. Further details are included in the Supplementary Material.

\noindent
\textbf{Results and discussion.}  The results of this first experiment are reported in~\cref{fig:q1}.
The behavior of both competitors on dSprites and MPI3D was extremely stable, owing to the fact that these data sets cover an essentially exhaustive set of variations for all generative factors, so we report their hold-out performance on the test set.
Since for CelebA variance was non-negligible, we ran both methods $7$ times varying the random seed used to initialize the network and report the average performance across runs and its standard deviation.

In addition to alignment, we also report explicitness \citep{eastwood2018framework}, which measures how well the linear regressor employed by DCI fits the generative factors. The higher, the better. Details on its evaluation are included in Suppl. Material.

The plots clearly show that, although the two methods achieve high and comparable accuracy in all settings, \methods attain better alignment in all data sets and for all supervision regimes than CBNMs, with a single exception in CelebA using low values of supervision, for a total of $13$ wins out of $15$ cases.
In all $disentanglement$ data sets, there is a clear margin between the alignment achieved by \methods and that of CBNMs: performances vary up to maximum of 15\% in dSprites, and a minumum of 8\% in MPI3D. 
In CelebA, the gap is evident with full supervision (almost 8\% of difference in alignment), and \methods still attain overall better scores in the 25\% and 50\% regime. On the other hand, performance are lower, but comparable, with 10\% supervision. The case at 1\% refers to an extreme situation where both CBNMs and \methods struggle to align with generative factors, as is clear also from the very low explicitness.   

In dSprites and MPI3D, both \methods and CBNMs quickly achieve very high alignment at $1\%$ supervision, as expected \cite{locatello2020disentangling}, whereas better results in CelebA are obtained with growing supervision.
Also, both models display similar stability on this data set, as shown by the error bars in the plot.

\subsection{\methods are leak-proof}
\label{sec:q3-leakage}

Next, we evaluated robustness to concept leakage in two scenarios that differ in whether the unobserved generative factors are disentangled with the observed ones or not, see~\cref{sec:alignment-and-leakage}.  In both experiments, we compare \methods with a CBNM and a modified \method where the OSR component has been removed (denoted CG-VAE).

\noindent
\textbf{Leakage due to unobserved entangled factors.}  We start by replicating the experiment of~\citet{mahinpei2021promises}:  {the goal is to discriminate between even and odd MNIST images using a latent representation $\vZ = (Z_4, Z_5)$ obtained by training (with complete supervision on the generative factors) \textit{only} on examples of $4$'s and $5$'s. Leakage occurs if the learned representation can be used to solve the prediction task better than random on a test set where all digits except $4$ and $5$ occur.}\footnote{\citet{margeloiu2021concept} perform classification using a multi-layer perceptron on top of $\vz$.  Following the CBM literature, we use a linear classifier instead.  Leakage occurs regardless.}  During training, we use the digit label for conditioning the prior $p(\vZ\mid\vY)$ of the \method. {More qualitative results are collected in \cref{sec: concept leakage mnist}.}

\begin{figure}[!h]
    \centering
    \begin{tabular}{ccc}
        \includegraphics[height=8em]{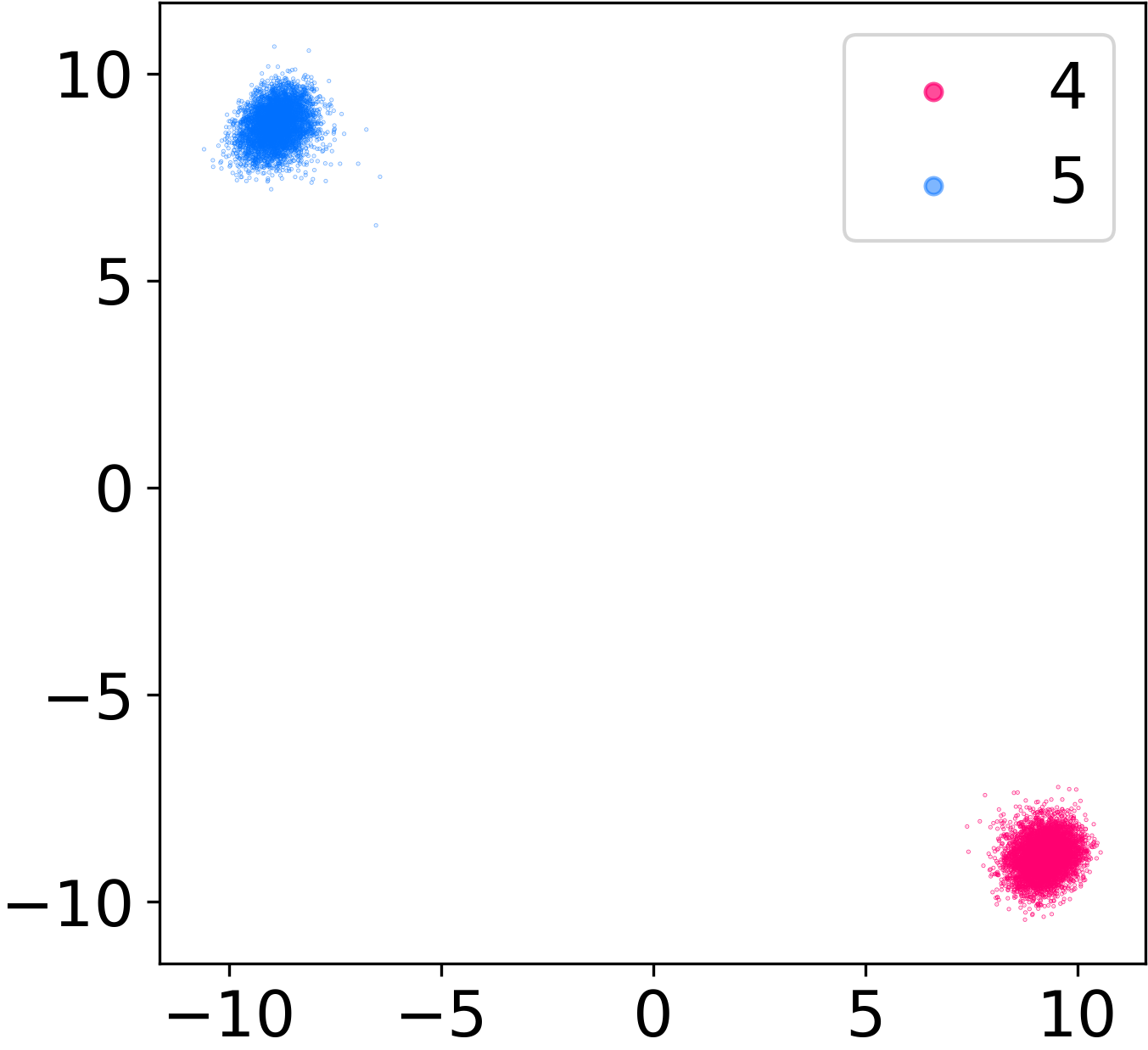} &
        \includegraphics[height=8em]{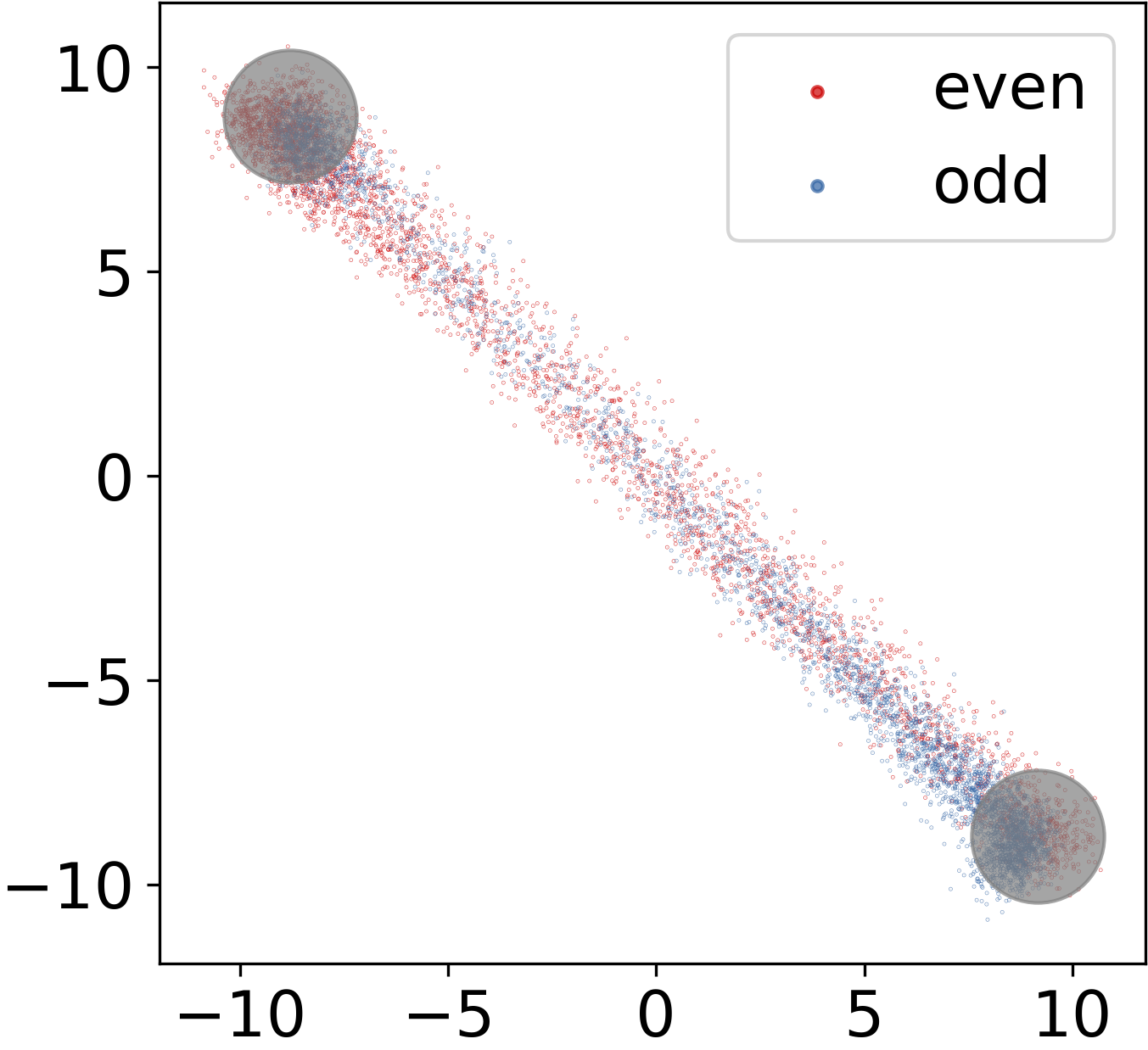} &
        \includegraphics[height=8em]{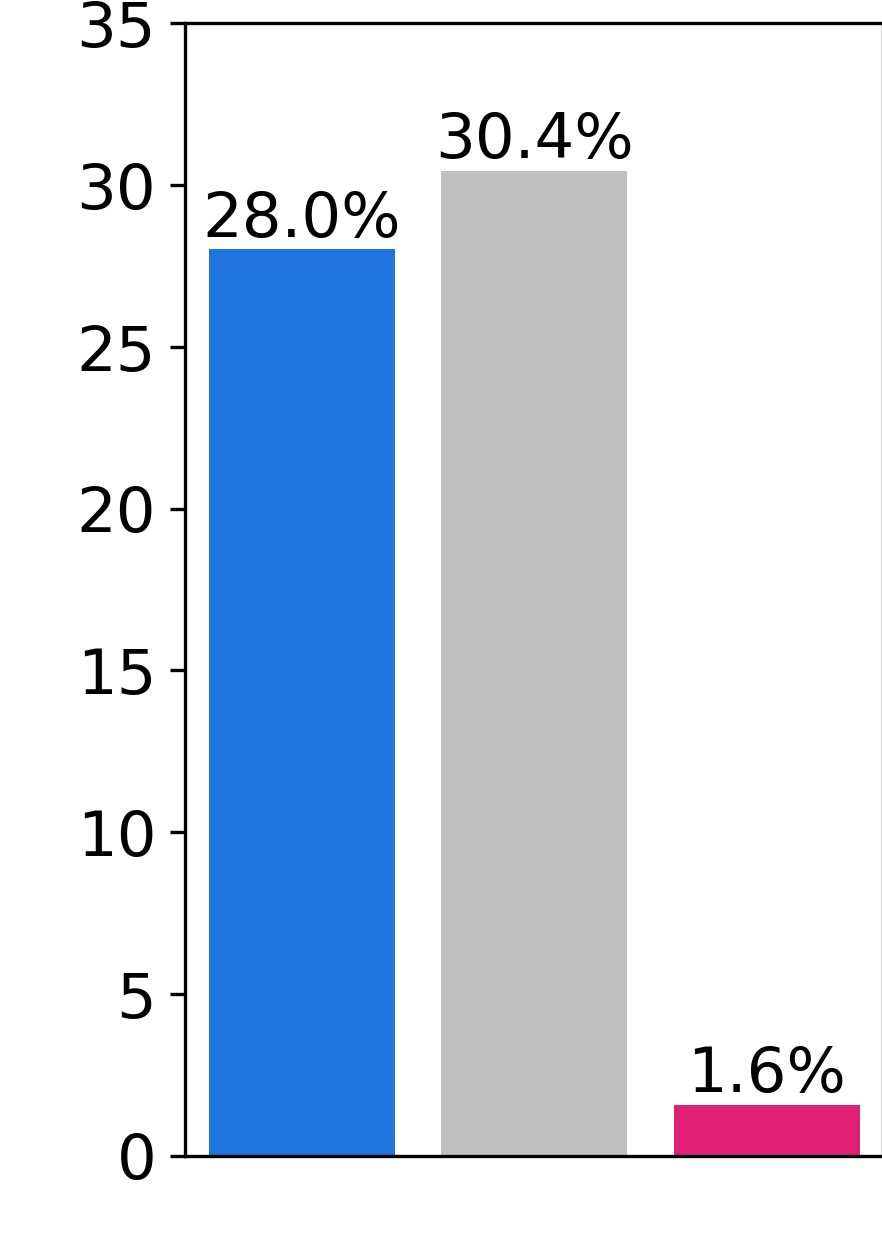}
        \\
        (a) & (b) & (c)
    \end{tabular}
    \caption{
    \textbf{\methods are leak-proof on MNIST.}
    (\textit{a}) Training set embedded by \method with $\beta = 100$; axes indicate $z_4$ and $z_5$ and color the concept label, \ie \textcolor{glancecolor}{$4$} vs.\@ \textcolor{cbnmcolor}{$5$}.
    (\textit{b}) Latent representations of the test images, divided in \textcolor{red}{even} vs.\@ \textcolor{blue}{odd}. Every ball in light gray denotes the region $ | \mu_y - \vz | < \eta_y$ for each class prototype $y$. For more details, refer to~\cref{sec:alignment-and-leakage}.
    (\textit{c}) Information Leakage performances of the considered models: \textcolor{blue}{CBNM}, \textcolor{darkgray}{CG-VAE} and \textcolor{red}{\method}.
    }
    \label{fig:architecture mnist}
\end{figure}

\cref{fig:architecture mnist} (a, b) illustrates the latent representations of the training and test set output by a \method:  since the two digits are mutually exclusive, the model has learned to map all instances along the $(z_4, z_5)$ diagonal. {This is where OSR kicks in:  if an input is identified as open-set, $T$ is predicted as $0$ by the OSR component and the input is rejected.}   In all leakage experiments, we implement rejection by predicting a random label.
Since MNIST is balanced, we measure leakage by computing the difference in accuracy between the classifier and an ideal random predictor, \ie $2 \cdot |\mathrm{acc} - \frac{1}{2}|$:  the smaller, the better.
The results, shown in \cref{fig:architecture mnist} (c), show a substantial difference between \method and the other approaches.
Consistently with the values reported in~\citep{mahinpei2021promises}, CBNMs are affected by a considerable amount of leakage, around $28\%$.
This is not the case for our \method:  most (approx. $85\%$) test images are correctly identified as open-set and rejected, leading to a very low (about $2\%$) leakage, $26\%$ less than CBNMs.
The results for CG-VAE also indicate that removing the open-set component from \methods dramatically increases leakage back to around $30\%$.  This shows that alignment and disentanglement alone are not sufficient, and that the open-set component plays a critical role for preventing leakage.

\begin{figure}[!h]
    \centering
    \begin{tabular}{cccc}
        \includegraphics[height=8em]{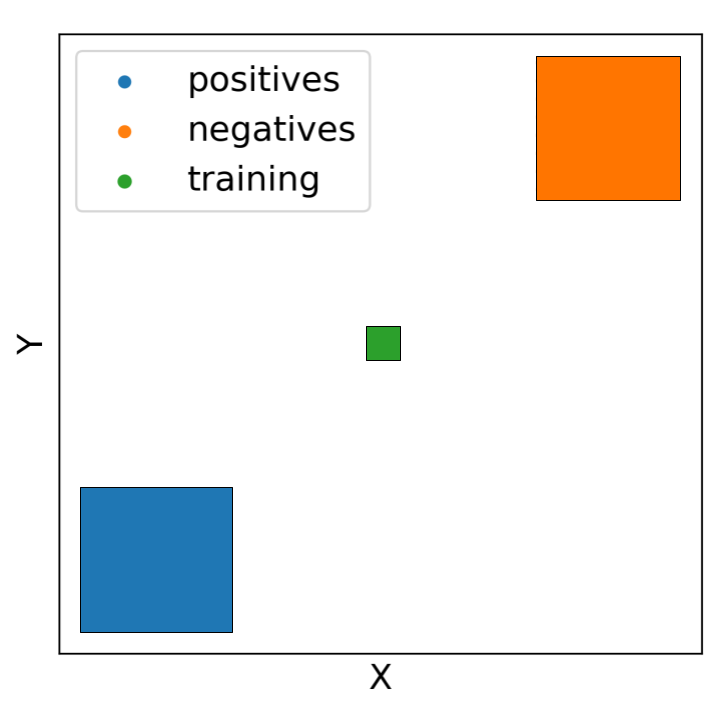} &
        \includegraphics[height=8em]{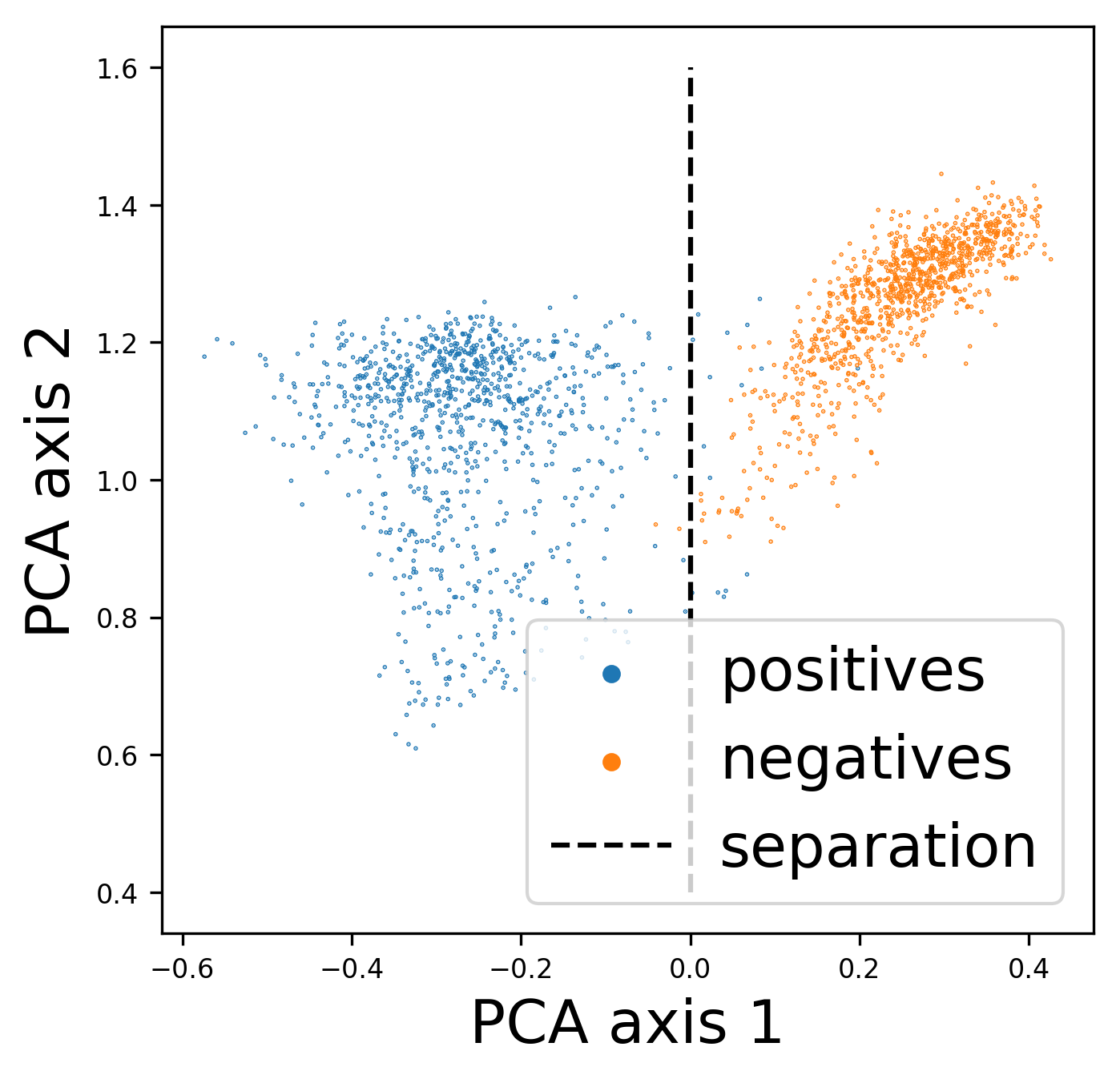} &
        \includegraphics[height=8em]{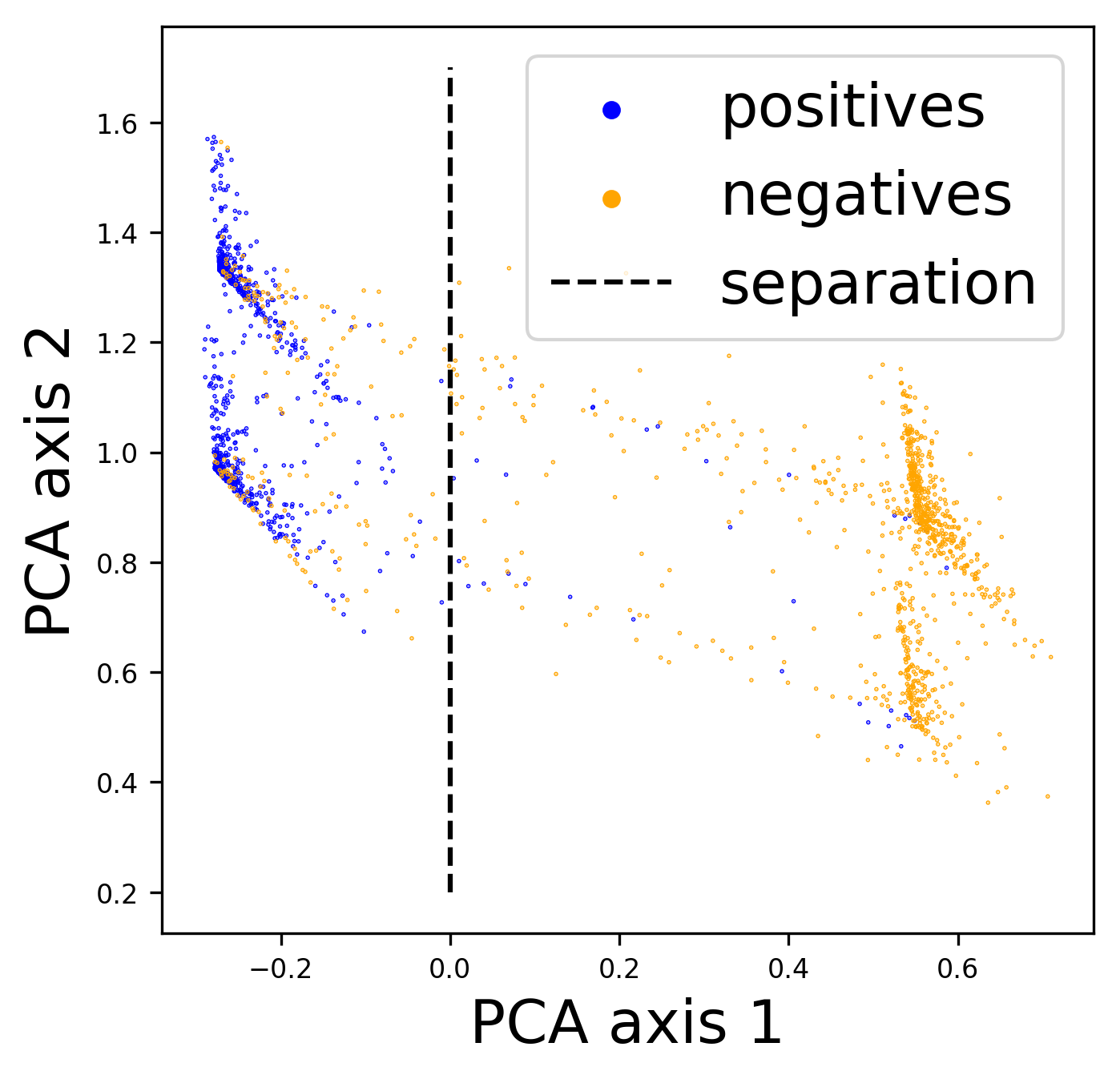} &
        \includegraphics[height=8em]{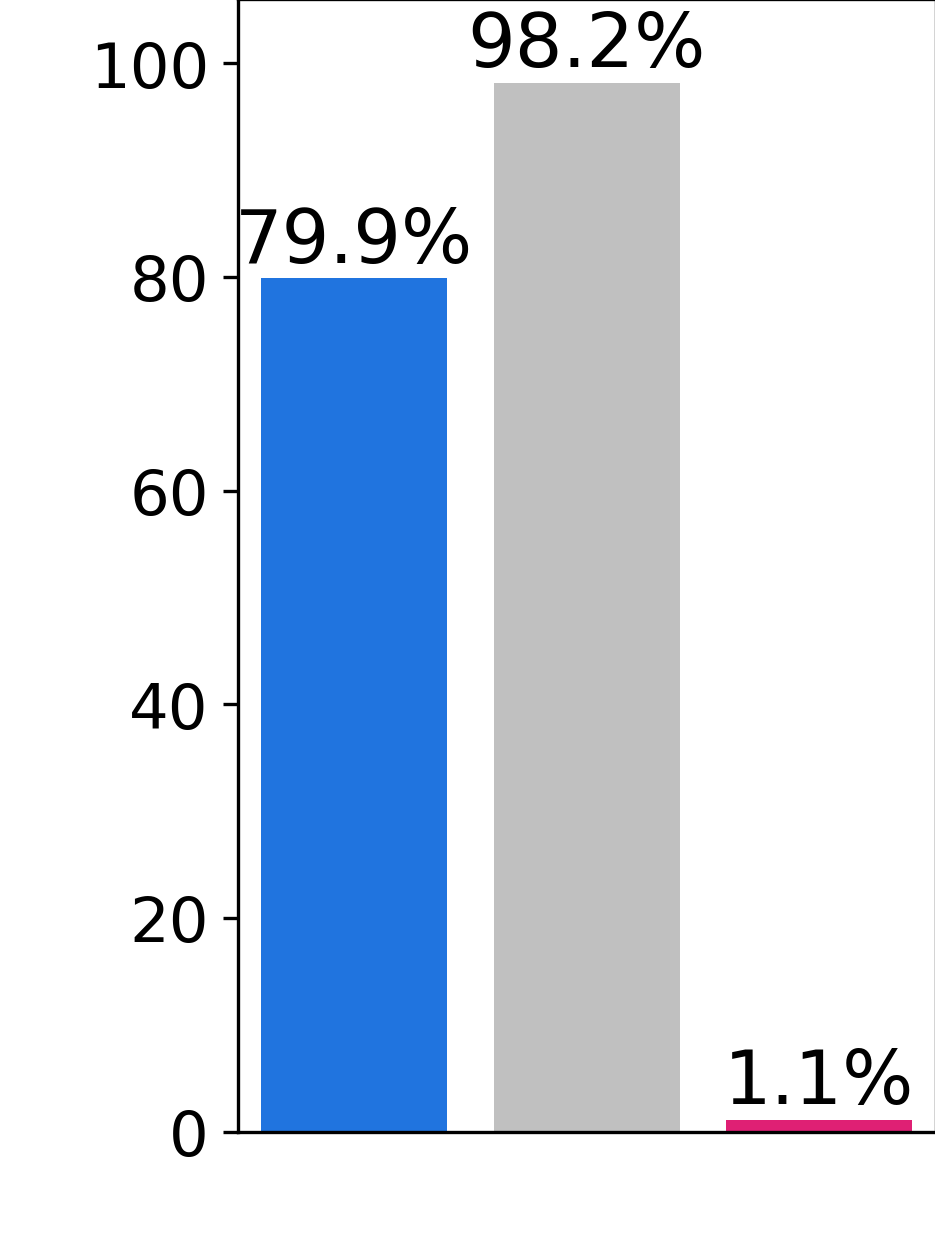}
        \\
        (a) & (b) & (c) & (d)
    \end{tabular}
    \caption{\textbf{\methods are leak-proof on dSprites.}
    (\textit{a}) The variations over \texttt{pos\_x} and \texttt{pos\_y} for the \textcolor{ForestGreen}{training} set, and for the test set, divided in \textcolor{blue}{positives} vs.\@ \textcolor{orange}{negatives}.
    (\textit{b}) PCA reduction for \method over the its five latent factors.
    (\textit{c}) PCA reduction for CBNM; the dotted line indicates the separating hyperplane predicted in the second phase.
    (\textit{d}) Leakage \% for \textcolor{blue}{CBNM}, \textcolor{darkgray}{CG-VAE} and \textcolor{red}{\method}.
    }
    \label{fig:architecture dsprites}
\end{figure}

\noindent
\textbf{Leakage due to unobserved disentangled factors.}  Next, we analyze concept leakage between \textit{disentangled} generative factors using the dSprites data set.
To this end, we defined a binary classification task in which the ground-truth label depends on \texttt{position\_x} and \texttt{position\_y} only.  In particular, instances within a fixed distance from $(0, 0)$ are annotated as positive and the rest as negative, as shown in~\cref{fig:architecture dsprites} (a).
In order to trigger leakage, all competitors are trained (using full concept-level supervision, as before) on training images where \texttt{shape}, \texttt{size} and \texttt{rotation} vary, but \texttt{position\_x} and \texttt{position\_y} are almost constant (they range in a small interval around $(0.5, 0.5)$, cf.~\cref{fig:architecture dsprites}).
leakage occurs if the learned model can successfully classify test instances where \texttt{position\_x} and \texttt{position\_y} are no longer fixed. {More qualitative results can be found in \cref{sec: concept leakage dsprites}.}

For both competitors, we encode \texttt{shape} using a 3D one-hot encoding and \texttt{size} and \texttt{rotation} as continuous variables.  During training, we use the $\texttt{shape}$ annotation for conditioning the prior $p(\vZ\mid\vY)$ of the \method.  The first two PCA components of the latent representations acquired by our \method and by a CBNM are shown, rotated so as to be separable on the first axis, in~\cref{fig:architecture dsprites} (b, c):  in both cases, it is possible to separate positives from negatives based on the obtained representations in the five latent dimensions.  As shown in~\cref{fig:architecture dsprites} (d), this means that both CBNM and CG-VAE suffer from very large leakage, $80\%$ and $98\%$, respectively.  In contrast, OSR allows us to correctly identify and reject almost all test instances, leading to negligible leakage even in this disentangled setting.

\section{Related Work}

\noindent
\textbf{Concept-based explainability.}  Concepts lie at the heart of AI~\citep{muggleton1994inductive} and have recently resurfaced as a natural medium for communicating with human stakeholders~\citep{kambhampati2021symbols}.
In explainable AI, this was first exploited by approaches like TCAV~\citep{kim2018interpretability}, which extract local concept-based explanations from black-box models using concept-level supervision to define the target concepts.  Post-hoc explanations, however, are notoriously unfaithful to the model's reasoning~\citep{dombrowski2019explanations,teso2019toward,sixt2020explanations}.
CBMs, including \methods, avoid this issue by leveraging concept-like representations directly for computing their predictions.
Existing CBMs model concepts using prototypes~\citep{li2018deep,chen2019looks,rymarczyk2020protopshare,nauta2021neural} or other representations~\citep{alvarez2018towards,koh2020concept,losch2019interpretability,chen2020concept}, but they seek interpretability using heuristics, and the quality of concepts they acquire has been called into question~\citep{nauta2021looks,hoffmann2021looks,mahinpei2021promises,margeloiu2021concept}.   We show that disentangled representation learning helps in this regard.

\noindent
\textbf{Disentanglement and interpretability.}  Interpretability is one of the main driving factors behind the development of disentangled representation learning~\citep{bengio2013representation,kulkarni2015deep,chen2016infogan}.
These approaches however make no distinction between interpretable and non-interpretable generative factors and generally focus on properties \textit{of the world}, like independence between causal mechanisms~\citep{scholkopf2021toward} or invariances~\citep{higgins2016beta}.  Interpretability, however, depends on human factors that are not well understood and therefore usually ignored~\citep{lipton2018mythos,miller2019explanation}.  
The link between disentanglement and interpretability has never been made explicit.
Importantly, in contrast to alignment, disentanglement does not require that the map between matching generative and learned factors preserves semantics. We remark that other VAE-based classifiers either do not tackle disentanglement or are unconcerned with concept leakage~\citep{Steenkiste2019AreDR, Xu2016SemisupervisedVA, sun2020conditional}.

\noindent
\textbf{Disentanglement and CBMs.}  Neither the literature on disentanglement nor the one on CBMs have attempted to formalize the notion of interpretability or to establish a proper link between the latter and disentanglement.
The work of \citet{kazhdan2021disentanglement} is the only one to compare techniques for disentangled representation learning and concept acquisition, however it makes no attempt at linking the two notions.  Our work fills this gap.

\begin{ack}
The research of ST and AP was partially supported by TAILOR, a project funded by EU Horizon 2020 research and innovation programme under GA No 952215. 
\end{ack}

\bibliographystyle{unsrtnat}
\bibliography{bibliography,explanatory-supervision}

\begin{thebibliography}{70}
\providecommand{\natexlab}[1]{#1}
\providecommand{\url}[1]{\texttt{#1}}
\expandafter\ifx\csname urlstyle\endcsname\relax
  \providecommand{\doi}[1]{doi: #1}\else
  \providecommand{\doi}{doi: \begingroup \urlstyle{rm}\Url}\fi

\bibitem[Alvarez-Melis and Jaakkola(2018)]{alvarez2018towards}
David Alvarez-Melis and Tommi~S Jaakkola.
\newblock Towards robust interpretability with self-explaining neural networks.
\newblock In \emph{Proceedings of the 32nd International Conference on Neural
  Information Processing Systems}, pages 7786--7795, 2018.

\bibitem[Li et~al.(2018)Li, Liu, Chen, and Rudin]{li2018deep}
Oscar Li, Hao Liu, Chaofan Chen, and Cynthia Rudin.
\newblock Deep learning for case-based reasoning through prototypes: A neural
  network that explains its predictions.
\newblock In \emph{Proceedings of the AAAI Conference on Artificial
  Intelligence}, 2018.

\bibitem[Chen et~al.(2019)Chen, Li, Tao, Barnett, Rudin, and Su]{chen2019looks}
Chaofan Chen, Oscar Li, Daniel Tao, Alina Barnett, Cynthia Rudin, and
  Jonathan~K Su.
\newblock This looks like that: Deep learning for interpretable image
  recognition.
\newblock \emph{Advances in Neural Information Processing Systems},
  32:\penalty0 8930--8941, 2019.

\bibitem[Losch et~al.(2019)Losch, Fritz, and
  Schiele]{losch2019interpretability}
Max Losch, Mario Fritz, and Bernt Schiele.
\newblock {Interpretability beyond classification output: Semantic Bottleneck
  Networks}.
\newblock \emph{arXiv preprint arXiv:1907.10882}, 2019.

\bibitem[Chen et~al.(2020)Chen, Bei, and Rudin]{chen2020concept}
Zhi Chen, Yijie Bei, and Cynthia Rudin.
\newblock Concept whitening for interpretable image recognition.
\newblock \emph{Nature Machine Intelligence}, 2\penalty0 (12):\penalty0
  772--782, 2020.

\bibitem[Rudin(2019)]{rudin2019stop}
Cynthia Rudin.
\newblock Stop explaining black box machine learning models for high stakes
  decisions and use interpretable models instead.
\newblock \emph{Nature Machine Intelligence}, 1\penalty0 (5):\penalty0
  206--215, 2019.

\bibitem[Mahinpei et~al.(2021)Mahinpei, Clark, Lage, Doshi-Velez, and
  Pan]{mahinpei2021promises}
Anita Mahinpei, Justin Clark, Isaac Lage, Finale Doshi-Velez, and Weiwei Pan.
\newblock Promises and pitfalls of black-box concept learning models.
\newblock In \emph{International Conference on Machine Learning: Workshop on
  Theoretic Foundation, Criticism, and Application Trend of Explainable AI},
  volume~1, pages 1--13, 2021.

\bibitem[Margeloiu et~al.(2021)Margeloiu, Ashman, Bhatt, Chen, Jamnik, and
  Weller]{margeloiu2021concept}
Andrei Margeloiu, Matthew Ashman, Umang Bhatt, Yanzhi Chen, Mateja Jamnik, and
  Adrian Weller.
\newblock Do concept bottleneck models learn as intended?
\newblock \emph{arXiv preprint arXiv:2105.04289}, 2021.

\bibitem[Sch{\"o}lkopf et~al.(2021)Sch{\"o}lkopf, Locatello, Bauer, Ke,
  Kalchbrenner, Goyal, and Bengio]{scholkopf2021toward}
Bernhard Sch{\"o}lkopf, Francesco Locatello, Stefan Bauer, Nan~Rosemary Ke, Nal
  Kalchbrenner, Anirudh Goyal, and Yoshua Bengio.
\newblock Toward causal representation learning.
\newblock \emph{Proceedings of the IEEE}, 109\penalty0 (5):\penalty0 612--634,
  2021.

\bibitem[Scheirer et~al.(2012)Scheirer, de~Rezende~Rocha, Sapkota, and
  Boult]{scheirer2012toward}
Walter~J Scheirer, Anderson de~Rezende~Rocha, Archana Sapkota, and Terrance~E
  Boult.
\newblock Toward open set recognition.
\newblock \emph{IEEE transactions on pattern analysis and machine
  intelligence}, 35\penalty0 (7):\penalty0 1757--1772, 2012.

\bibitem[Kambhampati et~al.(2022)Kambhampati, Sreedharan, Verma, Zha, and
  Guan]{kambhampati2021symbols}
Subbarao Kambhampati, Sarath Sreedharan, Mudit Verma, Yantian Zha, and Lin
  Guan.
\newblock {Symbols as a Lingua Franca for Bridging Human-AI Chasm for
  Explainable and Advisable AI Systems}.
\newblock In \emph{Proceedings of Thirty-Sixth AAAI Conference on Artificial
  Intelligence (AAAI)}, 2022.

\bibitem[Lipton(2018)]{lipton2018mythos}
Zachary~C Lipton.
\newblock The mythos of model interpretability: In machine learning, the
  concept of interpretability is both important and slippery.
\newblock \emph{Queue}, 16\penalty0 (3):\penalty0 31--57, 2018.

\bibitem[Hase and Bansal(2020)]{hase2020evaluating}
Peter Hase and Mohit Bansal.
\newblock {Evaluating Explainable AI: Which Algorithmic Explanations Help Users
  Predict Model Behavior?}
\newblock In \emph{Proceedings of the 58th Annual Meeting of the Association
  for Computational Linguistics}, pages 5540--5552, 2020.

\bibitem[Guidotti et~al.(2018)Guidotti, Monreale, Ruggieri, Turini, Giannotti,
  and Pedreschi]{guidotti2018survey}
Riccardo Guidotti, Anna Monreale, Salvatore Ruggieri, Franco Turini, Fosca
  Giannotti, and Dino Pedreschi.
\newblock A survey of methods for explaining black box models.
\newblock \emph{ACM computing surveys (CSUR)}, 51\penalty0 (5):\penalty0 1--42,
  2018.

\bibitem[Kim et~al.(2016)Kim, Khanna, and Koyejo]{kim2016examples}
Been Kim, Rajiv Khanna, and Oluwasanmi~O Koyejo.
\newblock Examples are not enough, learn to criticize! criticism for
  interpretability.
\newblock \emph{Advances in neural information processing systems}, 29, 2016.

\bibitem[Rymarczyk et~al.(2021)Rymarczyk, Struski, Tabor, and
  Zieli\'{n}ski]{rymarczyk2020protopshare}
Dawid Rymarczyk, \L{}ukasz Struski, Jacek Tabor, and Bartosz Zieli\'{n}ski.
\newblock {ProtoPShare: Prototypical Parts Sharing for Similarity Discovery in
  Interpretable Image Classification}.
\newblock In \emph{Proceedings of the 27th ACM SIGKDD Conference on Knowledge
  Discovery \& Data Mining}, page 1420–1430, 2021.

\bibitem[Nauta et~al.(2021{\natexlab{a}})Nauta, van Bree, and
  Seifert]{nauta2021neural}
Meike Nauta, Ron van Bree, and Christin Seifert.
\newblock Neural prototype trees for interpretable fine-grained image
  recognition.
\newblock In \emph{Proceedings of the IEEE/CVF Conference on Computer Vision
  and Pattern Recognition}, pages 14933--14943, 2021{\natexlab{a}}.

\bibitem[Singh and Yow(2021)]{singh2021these}
Gurmail Singh and Kin-Choong Yow.
\newblock These do not look like those: An interpretable deep learning model
  for image recognition.
\newblock \emph{IEEE Access}, 9:\penalty0 41482--41493, 2021.

\bibitem[Hoffmann et~al.(2021)Hoffmann, Fanconi, Rade, and
  Kohler]{hoffmann2021looks}
Adrian Hoffmann, Claudio Fanconi, Rahul Rade, and Jonas Kohler.
\newblock This looks like that... does it? shortcomings of latent space
  prototype interpretability in deep networks.
\newblock \emph{arXiv preprint arXiv:2105.02968}, 2021.

\bibitem[Koh et~al.(2020)Koh, Nguyen, Tang, Mussmann, Pierson, Kim, and
  Liang]{koh2020concept}
Pang~Wei Koh, Thao Nguyen, Yew~Siang Tang, Stephen Mussmann, Emma Pierson, Been
  Kim, and Percy Liang.
\newblock Concept bottleneck models.
\newblock In \emph{International Conference on Machine Learning}, pages
  5338--5348. PMLR, 2020.

\bibitem[Deng et~al.(2009)Deng, Dong, Socher, Li, Li, and
  Fei-Fei]{deng2009imagenet}
Jia Deng, Wei Dong, Richard Socher, Li-Jia Li, Kai Li, and Li~Fei-Fei.
\newblock Imagenet: A large-scale hierarchical image database.
\newblock In \emph{2009 IEEE conference on computer vision and pattern
  recognition}, pages 248--255. Ieee, 2009.

\bibitem[Sch{\"o}lkopf et~al.(2012)Sch{\"o}lkopf, Janzing, Peters, Sgouritsa,
  Zhang, and Mooij]{scholkopf2012causal}
Bernhard Sch{\"o}lkopf, Dominik Janzing, Jonas Peters, Eleni Sgouritsa, Kun
  Zhang, and Joris Mooij.
\newblock On causal and anticausal learning.
\newblock \emph{arXiv preprint arXiv:1206.6471}, 2012.

\bibitem[Suter et~al.(2019)Suter, Miladinovic, Sch{\"o}lkopf, and
  Bauer]{suter2019robustly}
Raphael Suter, Djordje Miladinovic, Bernhard Sch{\"o}lkopf, and Stefan Bauer.
\newblock Robustly disentangled causal mechanisms: Validating deep
  representations for interventional robustness.
\newblock In \emph{International Conference on Machine Learning}, pages
  6056--6065. PMLR, 2019.

\bibitem[Peters et~al.(2017)Peters, Janzing, and
  Sch{\"o}lkopf]{peters2017elements}
Jonas Peters, Dominik Janzing, and Bernhard Sch{\"o}lkopf.
\newblock \emph{Elements of causal inference: foundations and learning
  algorithms}.
\newblock 2017.

\bibitem[Reddy et~al.(2022)Reddy, Benin~Godfrey, and
  Balasubramanian]{reddy2021causally}
Abbavaram~Gowtham Reddy, L~Benin~Godfrey, and Vineeth~N Balasubramanian.
\newblock On causally disentangled representations.
\newblock In \emph{Proceedings of the AAAI Conference on Artificial
  Intelligence}, 2022.

\bibitem[Gabbay et~al.(2021)Gabbay, Cohen, and Hoshen]{gabbay2021image}
Aviv Gabbay, Niv Cohen, and Yedid Hoshen.
\newblock An image is worth more than a thousand words: Towards disentanglement
  in the wild.
\newblock \emph{Advances in Neural Information Processing Systems}, 34, 2021.

\bibitem[Matthey et~al.(2017)Matthey, Higgins, Hassabis, and
  Lerchner]{dsprites17}
Loic Matthey, Irina Higgins, Demis Hassabis, and Alexander Lerchner.
\newblock dsprites: Disentanglement testing sprites dataset.
\newblock https://github.com/deepmind/dsprites-dataset/, 2017.

\bibitem[Liu et~al.(2015)Liu, Luo, Wang, and Tang]{liu2015faceattributes}
Ziwei Liu, Ping Luo, Xiaogang Wang, and Xiaoou Tang.
\newblock Deep learning face attributes in the wild.
\newblock In \emph{Proceedings of International Conference on Computer Vision
  (ICCV)}, December 2015.

\bibitem[Eastwood and Williams(2018)]{eastwood2018framework}
Cian Eastwood and Christopher~KI Williams.
\newblock A framework for the quantitative evaluation of disentangled
  representations.
\newblock In \emph{International Conference on Learning Representations}, 2018.

\bibitem[Zaidi et~al.(2020)Zaidi, Boilard, Gagnon, and
  Carbonneau]{zaidi2020measuring}
Julian Zaidi, Jonathan Boilard, Ghyslain Gagnon, and Marc-Andr{\'e} Carbonneau.
\newblock Measuring disentanglement: A review of metrics.
\newblock \emph{arXiv preprint arXiv:2012.09276}, 2020.

\bibitem[Locatello et~al.(2019)Locatello, Bauer, Lucic, Raetsch, Gelly,
  Sch{\"o}lkopf, and Bachem]{locatello2019challenging}
Francesco Locatello, Stefan Bauer, Mario Lucic, Gunnar Raetsch, Sylvain Gelly,
  Bernhard Sch{\"o}lkopf, and Olivier Bachem.
\newblock Challenging common assumptions in the unsupervised learning of
  disentangled representations.
\newblock In \emph{International Conference on Machine Learning}, pages
  4114--4124, 2019.

\bibitem[Locatello et~al.(2020{\natexlab{a}})Locatello, Tschannen, Bauer,
  R{\"a}tsch, Sch{\"o}lkopf, and Bachem]{locatello2020disentangling}
Francesco Locatello, Michael Tschannen, Stefan Bauer, Gunnar R{\"a}tsch,
  Bernhard Sch{\"o}lkopf, and Olivier Bachem.
\newblock Disentangling factors of variations using few labels.
\newblock In \emph{International Conference on Learning Representations},
  2020{\natexlab{a}}.

\bibitem[Khemakhem et~al.(2020)Khemakhem, Kingma, Monti, and
  Hyvarinen]{khemakhem2020variational}
Ilyes Khemakhem, Diederik Kingma, Ricardo Monti, and Aapo Hyvarinen.
\newblock Variational autoencoders and nonlinear ica: A unifying framework.
\newblock In \emph{International Conference on Artificial Intelligence and
  Statistics}, pages 2207--2217. PMLR, 2020.

\bibitem[Kingma and Welling(2014)]{kingma2014auto}
Diederik~P Kingma and Max Welling.
\newblock Auto-encoding variational bayes.
\newblock In \emph{International conference on machine learning}. PMLR, 2014.

\bibitem[Rezende et~al.(2014)Rezende, Mohamed, and
  Wierstra]{rezende2014stochastic}
Danilo~Jimenez Rezende, Shakir Mohamed, and Daan Wierstra.
\newblock Stochastic backpropagation and approximate inference in deep
  generative models.
\newblock In \emph{International conference on machine learning}. PMLR, 2014.

\bibitem[Sun et~al.(2020)Sun, Yang, Zhang, Ling, and Peng]{sun2020conditional}
Xin Sun, Zhenning Yang, Chi Zhang, Keck-Voon Ling, and Guohao Peng.
\newblock Conditional gaussian distribution learning for open set recognition.
\newblock In \emph{Proceedings of the IEEE/CVF Conference on Computer Vision
  and Pattern Recognition}, pages 13480--13489, 2020.

\bibitem[Kingma and Welling(2019)]{kingma2019introduction}
Diederik~P Kingma and Max Welling.
\newblock An introduction to variational autoencoders.
\newblock \emph{Foundations and Trends{\textregistered} in Machine Learning},
  12\penalty0 (4):\penalty0 307--392, 2019.

\bibitem[Esmaeili et~al.(2019)Esmaeili, Wu, Jain, Bozkurt, Siddharth, Paige,
  Brooks, Dy, and Meent]{esmaeili2019structured}
Babak Esmaeili, Hao Wu, Sarthak Jain, Alican Bozkurt, Narayanaswamy Siddharth,
  Brooks Paige, Dana~H Brooks, Jennifer Dy, and Jan-Willem Meent.
\newblock Structured disentangled representations.
\newblock In \emph{The 22nd International Conference on Artificial Intelligence
  and Statistics}, pages 2525--2534. PMLR, 2019.

\bibitem[Chen et~al.(2018)Chen, Li, Grosse, and Duvenaud]{chen2018isolating}
Ricky~TQ Chen, Xuechen Li, Roger Grosse, and David Duvenaud.
\newblock Isolating sources of disentanglement in vaes.
\newblock In \emph{Proceedings of the 32nd International Conference on Neural
  Information Processing Systems}, pages 2615--2625, 2018.

\bibitem[Zhao et~al.(2019)Zhao, Song, and Ermon]{Zhao_Song_Ermon_2019}
Shengjia Zhao, Jiaming Song, and Stefano Ermon.
\newblock Infovae: Balancing learning and inference in variational
  autoencoders.
\newblock In \emph{Proceedings of the aaai conference on artificial
  intelligence}, volume~33, pages 5885--5892, 2019.

\bibitem[Kumar et~al.(2018)Kumar, Sattigeri, and
  Balakrishnan]{kumar2018variational}
Abhishek Kumar, Prasanna Sattigeri, and Avinash Balakrishnan.
\newblock Variational inference of disentangled latent concepts from unlabeled
  observations.
\newblock In \emph{International Conference on Learning Representations}, 2018.

\bibitem[Rhodes and Lee(2021)]{rhodes2021local}
Travers Rhodes and Daniel Lee.
\newblock Local disentanglement in variational auto-encoders using jacobian $
  l\_1 $ regularization.
\newblock \emph{Advances in Neural Information Processing Systems},
  34:\penalty0 22708--22719, 2021.

\bibitem[Higgins et~al.(2016)Higgins, Matthey, Pal, Burgess, Glorot, Botvinick,
  Mohamed, and Lerchner]{higgins2016beta}
Irina Higgins, Loic Matthey, Arka Pal, Christopher Burgess, Xavier Glorot,
  Matthew Botvinick, Shakir Mohamed, and Alexander Lerchner.
\newblock $\beta$-vae: Learning basic visual concepts with a constrained
  variational framework.
\newblock In \emph{International Conference on Learning Representations}, 2016.

\bibitem[Bouchacourt et~al.(2018)Bouchacourt, Tomioka, and
  Nowozin]{bouchacourt2018multi}
Diane Bouchacourt, Ryota Tomioka, and Sebastian Nowozin.
\newblock Multi-level variational autoencoder: Learning disentangled
  representations from grouped observations.
\newblock In \emph{Thirty-Second AAAI Conference on Artificial Intelligence},
  2018.

\bibitem[Shu et~al.(2020)Shu, Chen, Kumar, Ermon, and Poole]{shu2020weakly}
Rui Shu, Yining Chen, Abhishek Kumar, Stefano Ermon, and Ben Poole.
\newblock Weakly supervised disentanglement with guarantees.
\newblock In \emph{International Conference on Learning Representations}, 2020.

\bibitem[Locatello et~al.(2020{\natexlab{b}})Locatello, Poole, R{\"a}tsch,
  Sch{\"o}lkopf, Bachem, and Tschannen]{locatello2020weakly}
Francesco Locatello, Ben Poole, Gunnar R{\"a}tsch, Bernhard Sch{\"o}lkopf,
  Olivier Bachem, and Michael Tschannen.
\newblock Weakly-supervised disentanglement without compromises.
\newblock In \emph{International Conference on Machine Learning}, pages
  6348--6359. PMLR, 2020{\natexlab{b}}.

\bibitem[Gabbay and Hoshen(2019)]{gabbay2019latent}
Aviv Gabbay and Yedid Hoshen.
\newblock Latent optimization for non-adversarial representation
  disentanglement.
\newblock \emph{arXiv preprint arXiv:1906.11796}, 2019.

\bibitem[Chen and Batmanghelich(2020)]{chen2020weakly}
Junxiang Chen and Kayhan Batmanghelich.
\newblock Weakly supervised disentanglement by pairwise similarities.
\newblock In \emph{Proceedings of the AAAI Conference on Artificial
  Intelligence}, volume~34, pages 3495--3502, 2020.

\bibitem[Stammer et~al.(2021)Stammer, Memmel, Schramowski, and
  Kersting]{stammer2021interactive}
Wolfgang Stammer, Marius Memmel, Patrick Schramowski, and Kristian Kersting.
\newblock Interactive disentanglement: Learning concepts by interacting with
  their prototype representations.
\newblock \emph{arXiv preprint arXiv:2112.02290}, 2021.

\bibitem[Ross and Doshi-Velez(2021)]{ross2021benchmarks}
Andrew Ross and Finale Doshi-Velez.
\newblock Benchmarks, algorithms, and metrics for hierarchical disentanglement.
\newblock In \emph{International Conference on Machine Learning}, pages
  9084--9094. PMLR, 2021.

\bibitem[Paszke et~al.(2019)Paszke, Gross, Massa, Lerer, Bradbury, Chanan,
  Killeen, Lin, Gimelshein, Antiga, et~al.]{paszke2019pytorch}
Adam Paszke, Sam Gross, Francisco Massa, Adam Lerer, James Bradbury, Gregory
  Chanan, Trevor Killeen, Zeming Lin, Natalia Gimelshein, Luca Antiga, et~al.
\newblock Pytorch: An imperative style, high-performance deep learning library.
\newblock \emph{Advances in neural information processing systems}, 32, 2019.

\bibitem[Abdi et~al.(2019)Abdi, Abolmaesumi, and Fels]{abdi2019variational}
Amir~H. Abdi, Purang Abolmaesumi, and Sidney Fels.
\newblock Variational learning with disentanglement-pytorch.
\newblock \emph{arXiv preprint arXiv:1912.05184}, 2019.

\bibitem[Gondal et~al.(2019)Gondal, Wuthrich, Miladinovic, Locatello, Breidt,
  Volchkov, Akpo, Bachem, Sch\"{o}lkopf, and Bauer]{NEURIPS2019_d97d404b}
Muhammad~Waleed Gondal, Manuel Wuthrich, Djordje Miladinovic, Francesco
  Locatello, Martin Breidt, Valentin Volchkov, Joel Akpo, Olivier Bachem,
  Bernhard Sch\"{o}lkopf, and Stefan Bauer.
\newblock On the transfer of inductive bias from simulation to the real world:
  a new disentanglement dataset.
\newblock In H.~Wallach, H.~Larochelle, A.~Beygelzimer, F.~d\textquotesingle
  Alch\'{e}-Buc, E.~Fox, and R.~Garnett, editors, \emph{Advances in Neural
  Information Processing Systems}, volume~32. Curran Associates, Inc., 2019.

\bibitem[Huang(1997)]{Huang97clusteringlarge}
Zhexue Huang.
\newblock Clustering large data sets with mixed numeric and categorical values.
\newblock In \emph{In The First Pacific-Asia Conference on Knowledge Discovery
  and Data Mining}, pages 21--34, 1997.

\bibitem[Ghosh et~al.(2019)Ghosh, Sajjadi, Vergari, Black, and
  Sch{\"o}lkopf]{ghosh2020deterministic}
Partha Ghosh, Mehdi~SM Sajjadi, Antonio Vergari, Michael Black, and Bernhard
  Sch{\"o}lkopf.
\newblock From variational to deterministic autoencoders.
\newblock \emph{arXiv preprint arXiv:1903.12436}, 2019.

\bibitem[Muggleton and De~Raedt(1994)]{muggleton1994inductive}
Stephen Muggleton and Luc De~Raedt.
\newblock Inductive logic programming: Theory and methods.
\newblock \emph{The Journal of Logic Programming}, 19:\penalty0 629--679, 1994.

\bibitem[Kim et~al.(2018)Kim, Wattenberg, Gilmer, Cai, Wexler, Viegas,
  et~al.]{kim2018interpretability}
Been Kim, Martin Wattenberg, Justin Gilmer, Carrie Cai, James Wexler, Fernanda
  Viegas, et~al.
\newblock Interpretability beyond feature attribution: Quantitative testing
  with concept activation vectors (tcav).
\newblock In \emph{International conference on machine learning}, pages
  2668--2677. PMLR, 2018.

\bibitem[Dombrowski et~al.(2019)Dombrowski, Alber, Anders, Ackermann,
  M{\"u}ller, and Kessel]{dombrowski2019explanations}
Ann-Kathrin Dombrowski, Maximillian Alber, Christopher Anders, Marcel
  Ackermann, Klaus-Robert M{\"u}ller, and Pan Kessel.
\newblock Explanations can be manipulated and geometry is to blame.
\newblock \emph{Advances in Neural Information Processing Systems},
  32:\penalty0 13589--13600, 2019.

\bibitem[Teso(2019)]{teso2019toward}
Stefano Teso.
\newblock Toward faithful explanatory active learning with self-explainable
  neural nets.
\newblock In \emph{Proceedings of the Workshop on Interactive Adaptive Learning
  (IAL 2019)}, pages 4--16, 2019.

\bibitem[Sixt et~al.(2020)Sixt, Granz, and Landgraf]{sixt2020explanations}
Leon Sixt, Maximilian Granz, and Tim Landgraf.
\newblock When explanations lie: Why many modified bp attributions fail.
\newblock In \emph{International Conference on Machine Learning}, pages
  9046--9057. PMLR, 2020.

\bibitem[Nauta et~al.(2021{\natexlab{b}})Nauta, Jutte, Provoost, and
  Seifert]{nauta2021looks}
Meike Nauta, Annemarie Jutte, Jesper Provoost, and Christin Seifert.
\newblock This looks like that, because... explaining prototypes for
  interpretable image recognition.
\newblock In \emph{Joint European Conference on Machine Learning and Knowledge
  Discovery in Databases}, pages 441--456. Springer, 2021{\natexlab{b}}.

\bibitem[Bengio et~al.(2013)Bengio, Courville, and
  Vincent]{bengio2013representation}
Yoshua Bengio, Aaron Courville, and Pascal Vincent.
\newblock Representation learning: A review and new perspectives.
\newblock \emph{IEEE transactions on pattern analysis and machine
  intelligence}, 35\penalty0 (8):\penalty0 1798--1828, 2013.

\bibitem[Kulkarni et~al.(2015)Kulkarni, Whitney, Kohli, and
  Tenenbaum]{kulkarni2015deep}
Tejas~D Kulkarni, William~F Whitney, Pushmeet Kohli, and Josh Tenenbaum.
\newblock Deep convolutional inverse graphics network.
\newblock \emph{Advances in neural information processing systems}, 28, 2015.

\bibitem[Chen et~al.(2016)Chen, Duan, Houthooft, Schulman, Sutskever, and
  Abbeel]{chen2016infogan}
Xi~Chen, Yan Duan, Rein Houthooft, John Schulman, Ilya Sutskever, and Pieter
  Abbeel.
\newblock {InfoGAN: Interpretable representation learning by information
  maximizing generative adversarial nets}.
\newblock \emph{Advances in neural information processing systems}, 29, 2016.

\bibitem[Miller(2019)]{miller2019explanation}
Tim Miller.
\newblock Explanation in artificial intelligence: Insights from the social
  sciences.
\newblock \emph{Artificial intelligence}, 267:\penalty0 1--38, 2019.

\bibitem[van Steenkiste et~al.(2019)van Steenkiste, Locatello, Schmidhuber, and
  Bachem]{Steenkiste2019AreDR}
Sjoerd van Steenkiste, Francesco Locatello, J{\"u}rgen Schmidhuber, and Olivier
  Bachem.
\newblock Are disentangled representations helpful for abstract visual
  reasoning?
\newblock In \emph{NeurIPS}, 2019.

\bibitem[Xu and Sun(2016)]{Xu2016SemisupervisedVA}
Weidi Xu and Haoze Sun.
\newblock Semi-supervised variational autoencoders for sequence classification.
\newblock \emph{ArXiv}, abs/1603.02514, 2016.

\bibitem[Kazhdan et~al.(2021)Kazhdan, Dimanov, Terre, Jamnik, Li{\`o}, and
  Weller]{kazhdan2021disentanglement}
Dmitry Kazhdan, Botty Dimanov, Helena~Andres Terre, Mateja Jamnik, Pietro
  Li{\`o}, and Adrian Weller.
\newblock Is disentanglement all you need? comparing concept-based \&
  disentanglement approaches.
\newblock \emph{arXiv preprint arXiv:2104.06917}, 2021.

\bibitem[Tolstikhin et~al.(2017)Tolstikhin, Bousquet, Gelly, and
  Schoelkopf]{tolstikhin2018wasserstein}
Ilya Tolstikhin, Olivier Bousquet, Sylvain Gelly, and Bernhard Schoelkopf.
\newblock Wasserstein auto-encoders.
\newblock \emph{arXiv preprint arXiv:1711.01558}, 2017.

\bibitem[Kingma()]{kingma15adam}
Adam Kingma.
\newblock Adam: a method for stochastic optimization, 3rd, int. conf. learn.
  represent. iclr 2015-conf.
\newblock \emph{Track Proc}, \penalty0 (1-15).

\end{thebibliography}

\newpage

\appendix

\section{Implementation details} \label{sec: implementation details}

\subsection{\method and CBNM Architectures}

In all experiments, we used exactly the same architecture and number of latent variables for both \methods and CBNMs to ensure a fair comparison.

\begin{table}[p]
    \centering
    \footnotesize
    \caption{Structure of the encoder network used for dSprites.}
    \begin{tabular}{cccc}
         \toprule
         \textsc{Input shape} & \textsc{Layer type} & \textsc{Parameters} & \textsc{Activation} \\ 
         \midrule
         $( 64, 64, 1 )$ & Convolution & {depth}=$32$,  {kernel}=$4$, {stride}=$2$, {padding}=$1$ & ReLU \\
          $(32,32, 32)$  & Convolution & depth=$32$,  kernel=$4$, stride=$2$, padding=$1$ & ReLU \\
          $(16, 16, 32)$  & Convolution & depth=$64$,  kernel=$4$, stride=$2$, padding=$1$ & ReLU \\
         $(8, 8, 64)$  & Convolution & depth=$128$,  kernel=$4$, stride=$2$, padding=$1$ & ReLU \\  
         $(4, 4, 128)$  & Convolution & depth=$256$,  kernel= $4$, stride=$2$, padding=$1$ & ReLU \\
         $(2, 2, 256)$ & Convolution & depth=$256$,  kernel=$4$, stride=$2$, padding=$1$ & ReLU \\
         $(1,1,256) $ & Flatten &   &  \\
         $(1,256)$ & Linear & dim=7+7, {bias} = True \\
         \bottomrule
    \end{tabular}
    
    \label{tab: dsprites-encoder}
\end{table}

\begin{table}[p]
    \centering
    \footnotesize
    \caption{Structure of the encoder network used for MPI3D.}
    \begin{tabular}{cccc}
         \toprule
         \textsc{Input shape} & \textsc{Layer type} & \textsc{Parameters} & \textsc{Activation} \\ 
         \midrule
         $( 64, 64, 3 )$ & Convolution & {depth}=$32$,  {kernel}=$3$, {stride}=$2$ & ReLU \\
          $(32,32, 32)$  & Convolution & depth=$32$,  kernel=$3$, stride=$2$ & ReLU \\
          $(16, 16, 32)$  & Convolution & depth=$64$,  kernel=$3$, stride=$2$ & ReLU \\
         $(8, 8, 64)$  & Convolution & depth=$64$,  kernel=$3$, stride=$2$ & ReLU \\  
         $(4, 4, 64)$  & Convolution & depth=$128$,  kernel= $3$, stride=$2$ & ReLU \\
         $(2, 2, 128)$ & Convolution & depth=$256$,  kernel=$3$, stride=$2$ & ReLU \\
         $(1,1,256) $ & Flatten &   &  \\
         $(1,256)$ & Linear & dim=21+21, {bias} = True \\
         \bottomrule
    \end{tabular}

    \label{tab: MPI3D-encoder}
\end{table}

\begin{table}[p]
    \centering
    \footnotesize
    \caption{Structure of the encoder network used for CelebA.}
    \begin{tabular}{ccccc}
         \toprule
         \textsc{Input shape} & \textsc{Layer type} & \textsc{Parameters} & \textsc{Filter} & \textsc{Activation} \\ 
         \midrule
         $( 64, 64, 3 )$ & Convolution & {depth}=$128$,  {kernel}=$5$, {stride}=$2$ & BatchNorm & ReLU \\
          $(30,30, 128)$  & Convolution & depth=$256$,  kernel=$5$, stride=$2$ & BatchNorm & ReLU \\
          $(13, 13, 256)$  & Convolution & depth=$512$,  kernel=$5$, stride=$2$ & BatchNorm & ReLU \\
         $(5,5,512)$  & Convolution & depth=$1028$,  kernel=$5$, stride=$2$ & BatchNorm & ReLU \\  
         $(1,1,1028)$  & Flatten &   &  \\
         $(1,1028)$ & Linear & dim=10+10, {bias} = True \\
         \bottomrule
    \end{tabular}

    \label{tab: CelebA-encoder}
\end{table}

\begin{table}[p]
    \centering
    \footnotesize
    \caption{Structure of the decoder network.}
    \begin{tabular}{ccccc}
         \toprule
         \textsc{Input shape} & \textsc{Layer type} & \textsc{Parameters} & \textsc{Activation} \\ 
         \midrule
         $(\text{dim}(\vz) )$ & Unsqueeze & & \\
         $(\text{dim}(\vz),1,1 )$ & Convolution & {depth}=$256$,  {kernel}=$1$, {stride}=$2$ & ReLU \\
          $(256,1,1)$  & Deconvolution & depth=$256$,  kernel=$4$, stride=$2$ & ReLU \\
          $(256,2,2)$  & Deconvolution & depth=$128$,  kernel=$4$, stride=$2$  & ReLU \\
         $(128,6,6)$  & Deconvolution & depth=$128$,  kernel=$4$, stride=$2$  & ReLU \\  
         $(128, 14, 14)$  & Deconvolution & depth=$64$,  kernel=$4$, stride=$2$  & ReLU \\
         $(64, 30, 30)$  & Deconvolution & depth=$64$,  kernel=$4$, stride=$2$  & ReLU \\
         $(64, 62, 62)$  & Deconvolution & depth=\texttt{num\_channels},  kernel=$4$, stride=$2$
         \\
        \bottomrule
    \end{tabular}

    \label{tab: decoder}
\end{table}

\noindent
\textbf{Encoder architectures:}
\begin{itemize}[leftmargin=1.25em]

    \item \textit{dSprites}:  We chose a rather standard architecture~\citep{abdi2019variational}.
    It comprises six 2D convolutional layers of depth $32$, $32$, $64$, $128$, $256$, and $256$, respectively, all with a kernel of size $4$, stride $2$, and padding $1$, and followed by ReLU activations.
    The output is flattened to a vector and passed through a dense layer to obtain the mean $\vmu(\vx)$ and (diagonal) variance $\vsigma(\vx)$ of the encoder distribution $\mathcal{N}(\vZ \mid \vmu (\vx), \mathrm{diag}(\vsigma(\vx)))$.

    \item \textit{MPI3D}:  We used the same architecture with slightly different convolutional depths of $32$, $32$, $64$, $64$, $128$, and $256$, changing also the kernel size to $3$ and removing padding, as per~\citep{abdi2019variational}. 

    \item \textit{CelebA}:  We leveraged the architecture of~\citet{ghosh2020deterministic}, which is a common reference for VAE models on CelebA-64~\citep{tolstikhin2018wasserstein}. The encoder is composed of four convolutions of depth $128$, $256$, $512$, $1024$ respectively, all with kernel size of $5$, stride of $2$, followed batch normalization and ReLU activation.

\end{itemize}
The models had exactly as many latent variables as generative factors for which supervision is available, which in our three data sets are $7$, $21$, and $10$, respectively.

\noindent
\textbf{Decoder architecture:}  All models share the same decoder architecture, obtained by stacking:
\begin{itemize}[leftmargin=1.25em]

    \item A 2D convolution on the latent space with a filter depth of 256, kernel size of 1, and stride of 2, followed by the ReLU activation;
    
    \item Five transposed 2D convolutions of depth $256$, $256$, $128$, $128$, $64$, $64$, and \texttt{num\_channels}, respectively, all with kernel of size 4 and stride 2.
    
\end{itemize}
Here, \texttt{num\_channels} is either 1 (dSprites) or 3 (MPI3D and CelebA).
The shape of the last layer was chosen so as to match the dimension of the input image.  Additional details can be found in the various Tables in this appendix.

\subsection{Supervision and Training}

\paragraph{Concept-level supervision.}  Depending on the supervision provided, only a fraction of the inputs was made available during training with their generative factors.
In dSprites and MPI3D all generative factors are matched by the models, whereas in the case of CelebA we restricted learning to those 10 attributes that are best fit by the CBNMs, namely: \texttt{bald}, \texttt{black hair}, \texttt{brown hair}, \texttt{blonde hair}, \texttt{eyeglasses}, \texttt{gray hair}, \texttt{male}, \texttt{no beard}, \texttt{smiling}, and  \texttt{wearing hat}. {Both CBNMs and \methods are jointly trained, meaning that optimization steps for the concepts and label supervision are taken simultaneously. Whenever concept supervision is lower than $100$\%, for those examples without concept annotations we trained both models using label supervision only. We did not evaluate other training strategies available for CBNMs (e.g., sequential training~\cite{koh2020concept}) as these appear to bring no benefit in terms of either performance nor leakage.}

\paragraph{Optimization setup.}  In all experiments, we used the Adam optimizer~\citep{kingma15adam} with default parameters $\beta_1 = 0.9$ and $\beta_2=0.999$.
For dSprites, we used a batch size of $64$ and fixed learning rate to $\eta = 4 \cdot 10^{-4}$,
while for MPI3D and CelebA we used a batch size of $100$ and annealed the learning rate from $10^{-7}$ to $\eta_{MPI} = 10^{-3}$ and $\eta_{CelebA} = 10^{-4}$, respectively.
To prevent overfitting, in CelebA we multiplied the learning rate by a factor of $0.95$ in each epoch and apply early stopping on the validation set, with a patience of 10 epochs.

Prior to training, we selected a reasonable value for the following hyper-parameters:
\begin{itemize}
    \item $\beta$: the weight of the KL divergence in Eq. \eqref{eq:elbo}.
    \item $\gamma$: the weight of the loss on the generative factors in Eq. \eqref{eq:elbo+int}.
    \item $\lambda$: the weight of the cross-entropy loss over the label, which is left implicit in~\cref{eq:elbo}.
\end{itemize}
For dSprites, we found a good balance for $\lambda = \gamma = 100$, while for MPI3D we achieved good performance with $\lambda = 10^3$ and $\gamma = 7 \cdot 10^3$. We adopted the same hyperparameters choice for CelebA, with the exception that we reduced the reconstruction error by $0.01$.
For all data sets, we cross-validated over different values of $\beta$ but we obtained better alignment performances with $\beta \approx 1$. This happens because we inject supervision on the latent factors (which is absent in regular $\beta$-VAEs~\cite{higgins2016beta}).

\subsection{Implementation of leakage tests}

\noindent
\textbf{MNIST.} For this dataset, we considered only Multi-Latyer Perceptrons instead of convolutions. Both the encoder and the decoder are composed by two linear layers with depth 128, and a dense layer connected to the latent space and to the input space, respectively. Further details are on~\cref{tab: all MNIST}. 

For the \method we considered a latent space of dimension 10 where the supervision on the 4 and 5 digits is used to fit the $\{z_4, z_5 \}$ latent factors. These two, constitute the latent subspace where leakage occurs, while the other are useful only for reconstruction. Conversely, for the CBNM we considered only two latent factors.  

During training of the latent encodings, we used stochastic gradient descent with learning rate $\eta = 0.001$, reducing it by 0.95 in each epoch for both CBNMs and \methods. The training was performed only on the 4 and 5 digits (in the usual training set partition for MNIST), for almost 50 epochs.
Afterwards, we considered the open-set representations, restricted to $\{z_4, z_5 \}$, as inputs for training a logistic regression for parity recognition. During the training, only the digits in the MNIST training set partition (exception made for 4 and 5) are considered, while performance are calculated on the test set.

\noindent
\textbf{dSprites.} We adopted the same architecture in the upper section, except that we reduced the latent space to 5 dimensions. As a reminder, during training all sprites are almost fixed at the center, therefore additional factors of variations for its position are needless. The training was performed over 300 epochs for both \methods and CBNMs, with $\eta=4 \cdot 10^{-4}$. After training, the representations of the open-set sprites (in which position is no longer fixed) are used to fit a logistic regression. In this case, the labels depend on whether the sprite is located at bottom-left corner or at the upper-right one, for more information refer to ~\cref{fig:architecture dsprites}. 

The classification performance was evaluated on a held-out test set for both models, under an 80/20 train/test split. 

\begin{table}[!h]
    \centering
    \caption{Encoder and Decoder structures for MNIST}
    \begin{tabular}{c|cccc}
         \hline 
         \textsc{Type} & \textsc{Input shape} & \textsc{Layer type} & \textsc{Parameters} & \textsc{Activation} \\ 
         \hline
         \textsc{Encoder} \\
         &$( 28,28 )$ & Flatten  &  & \\
         & $(784)$ & Linear & dim=$128$, bias=True  & ReLU  \\
         & $(128)$ & Linear &  dim=128, bias=True & ReLU \\
         & $(128)$ & Linear &  dim=10+10, bias=True &   \\
         \hline
        \textsc{Decoder} \\
         & $(\text{dim}(\vz))$ & Linear  & dim=$128$, bias=True  & ReLU  \\
          & $(128)$ & Linear  & dim=$128$, bias=True  & ReLU  \\
          & $(128)$ & Linear  & dim=$728$, bias=True  &   \\
         & $(728)$ & Unsqueeze &    \\
         \hline
 
    \end{tabular}

    \label{tab: all MNIST}
\end{table}

\section{DCI framework}
\label{sec:dci-framework}

In our case study, we are interested into DCI  in \cite{eastwood2018framework})  maps that linearly connect the $\vz 's$ to the $\vg 's$ . In order to evaluate alignment performances, the inverse map $\alpha^{-1} : \bbR^k \to \bbR^{n_I} $ is constructed from the latent space to the span of the $n_I$ generative factors. The latent representations and generative factors were normalized in the $[0,1]$ interval prior to learning.

\subsection{Alignment and explicitness}

The importance weights of this map are the absolute-values of the weights in the linear matrix of $\alpha^{-1}$, indicated as $B \in \bbR^{k \times n_I}$ in the main text. Then, the importance weights are used to evaluate the dispersion of the learned weights.  To this end, we measure each Shannon entropy $H_j$ on all $k$ latent factors:
\begin{equation}
H_j = -\sum_{i \in 1}^{n_I} \bar{b}_{ji} \log_n \bar{b}_{ji} \quad \text{where} \; \; \bar{b}_{ji} = b_{ji} \, \big / \sum_{\ell = 1}^{n_I} b_{j \ell} 
\end{equation}

Then, the average alignment is calculated as:
\begin{equation} \label{eq: definition alignment}
\text{alignment} = 1 - \sum_{j = 1}^k \rho_j H_j \quad \text{where} \; \;  \rho_j = \sum_{i=1}^{n_I} b_{ji} \, \Big / \sum_{j' = 1, i = 1}^{k, n_I} b_{j' i}
\end{equation}
and ranges in $[0, 1]$.
{Similarly, the quantity:
\begin{equation}
    \tilde{b}_{ji} = b_{ji} \, \big / \sum_{\ell = 1}^{k} b_{ \ell i}  \quad \mathrm{and} \quad \tilde{H}_i = \sum_{j=1}^k\tilde{b}_{ji} \log_k  \tilde{b}_{ji}
\end{equation}
is the \textit{completeness} of the latent representation, a measure akin to alignment (\cref{eq: definition alignment}) that quantities the degree to which each generative factor correlates with \textit{distinct} latent factors.  Alignment and completeness relate to different properties of the map: the higher the \textit{alignment}, the more each $Z_j$ depends on variations of only a single $G_i$. On the other hand, learning multiple $Z_j$'s capturing a single $G_i$ reduces the \textit{completeness}. As an illustrative example, consider the matrix:
$$ B = \begin{pmatrix} 
1 & 0 & 0 \\
0 & 0 & 1 \\
0 & 1 & 0 \\
0.2 & 0 & 0 \\
0 & 0.2 & 0 \\
\end{pmatrix}$$
From the above definitions, one gets $alignment =1$ and $completeness < 1$. This follows since each Shannon entropy for the alignment score is zero (as it is related to the rows), whereas the Shannon entropy for the completeness is greater than zero (it refers to the columns).  Moreover,  each latent variable $z_i$ depends only on the variations of a single generative factor $g_j$. }

We also calculate the explicitness of the map $\alpha$, which is related to the mean squared error (MSE) of the prediction. Since the MSE for random guessing for a variable in the $[0,1]$ interval is equal to 1/6, the explicitness becomes:
$$ \text{explicitness} = 1 - 6\cdot \text{MSE} $$

\subsection{Empirical evaluation}

For dSprites and MPI3D, all DCI quantities were calculated with the built-in evaluation code provided by \texttt{disentanglement\_lib}, \cite{locatello2019challenging}. For CelebA, since the 40 attributes in CelebA are not exhaustive for the image generation, we implemented computed DCI  as follows: \textit{(i)} we first converted the $J$ attributes $\vz_J$ and $\vg_J$ connected to \texttt{hair type} to a single concept $h$ and fit the model with Lasso regression to predict $g_h$ from $\vz$. Then,  \textit{(ii)} we trained a Logistic Regression with $l1$ penalty to predict the remaining $\vg 's$. Finally, we took both weights in \textit{(i)} and in \textit{(ii)} to compute the matrix $B \in \bbR^{6 \times 6}$. In this way, we determined alignment and explicitness for CelebA. We chose the lasso coefficient $\lambda = 0.01$ for both regressions.

{
\section{Open-Set Recognition Mechanism} \label{sec: openset appendix}

In this section, we provide additional details on the OSR mechanism introduced in \cref{sec:alignment-and-leakage}. 
Our method adapts the one of \citet{sun2020conditional}, which distinguishes between closed-set and open-set data points by combining a reconstruction check $\Gamma_r$ with a localization check $\Gamma_{ls}$.
The overall OSR check is given by:
\begin{equation}
\hat T = \Gamma_r \wedge \Gamma_{ls}
\end{equation}
After completing the training process, all the training instances are passed to the model to evaluate the thresholds:
\begin{itemize}

    \item The reconstruction threshold $\eta_r$ is the maximum real number such that a fixed percentage of training examples have reconstruction error less or equal to it.  At test time, given an instance $\vx$, let $\hat\eta = \norm{\vx - \hat{\vx}}^2$ be the reconstruction error. Then, $\Gamma_r=1$ (i.e., the check passes) if the empirical reconstruction error is less than the threshold, $\hat \eta < \eta_r$, otherwise $\Gamma_r= 0$.

    \item The latent-space distance thresholds are evaluated for each class-prototype embedded in the latent space $\mu_y = \bbE_{p(\vz | y)}[\vz] $. For each of them, we first evaluated the relative distance between point belonging to the class $y$ and the prototype $\mu_y$. Then, we evaluated a threshold $\eta_y$ on the distances, as to include a fixed percentage of training instances into the set ${\calB}_y = \{ \vz :\, \norm{\mu_y - \vz} < \eta_y  \}$. At test time, those points that do not belong to any set ${\calB}_y$ are predicted as open-set instances, i.e. $\Gamma_{ls} = 0$, otherwise $\Gamma_{ls} = 1$.

\end{itemize}
In our experiments, the threshold are obtained by fixing both reconstruction and latent space distance to keep the $95\%$ of training data. In the case of $\eta_y$, this quota has been reached singularly for each ${\calB}_y$, thus obtaining different values $\eta_y$'s from one another. Finally, combining both rejection methods we are sure the model would predict as closed-set at least the $90\%$ of training instances.}

\section{Concept Leakage in MNIST} \label{sec: concept leakage mnist}

{
We report here additional details for the concept leakage test on MNIST, which has been originally introduced by~\citet{margeloiu2021concept}. The experiment has two stages:
\begin{enumerate}

    \item At train time, the model is trained to align its representations to the concepts of $4$ and $5$, by passing full supervision on them. Both CBNMs and \methods are allotted two latent concepts, which we denote $(Z_4, Z_5)$. There is no downstream classification task in this stage.

    \item At test time, all MNIST images, excluding those of $4$'s and $5$'s, are encoded using the learned encoder and used to learn a classifier of even vs.\@ odd digits.  The performance of the resulting classifier, applied to non-$\{4, 5\}$ images, is then computed.

\end{enumerate}
In this experiment, concept leakage occurs if the accuracy on the downstream task is above the $50\%$.
}

\begin{figure}[!h]
    \centering
    \begin{tabular}{cccc}
    & \textsc{CBNM} & \textsc{VAE} &    \hspace{1em} \textsc{\method} \\
            \rotatebox{90}{\hspace{1.5em} \textsc{Closed set}} &
        \includegraphics[height=8em]{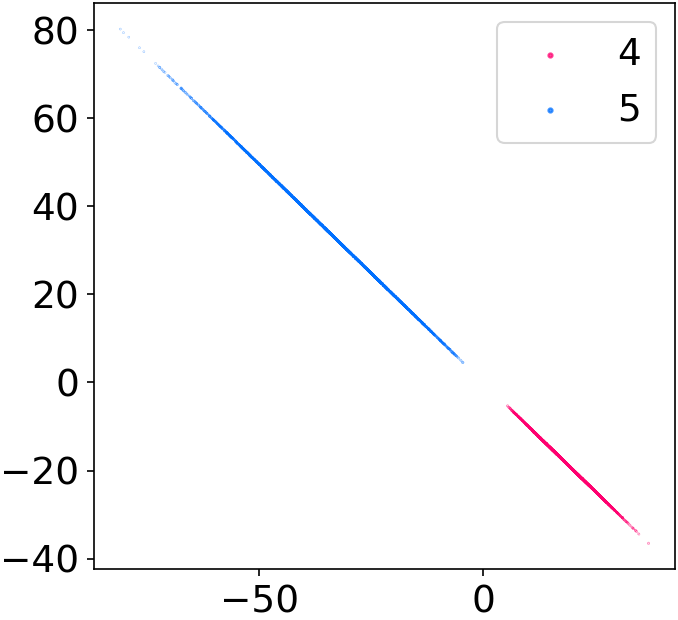} &
    \includegraphics[height=8em]{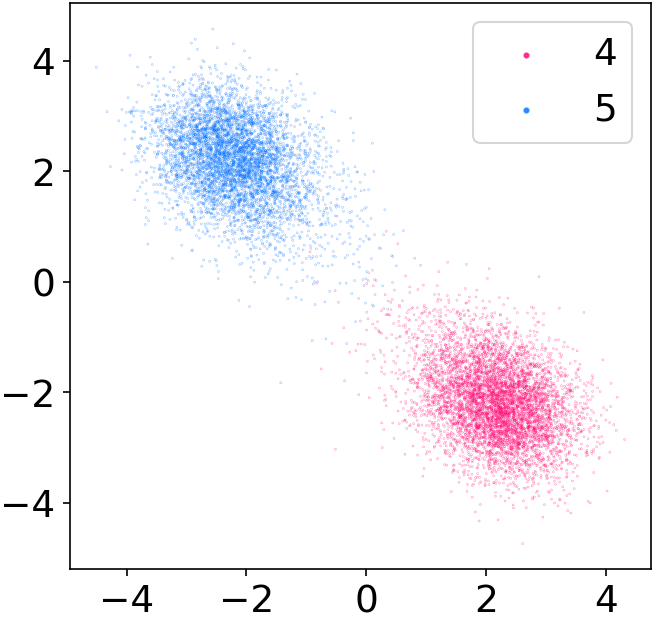}  &
    \includegraphics[height=8em]{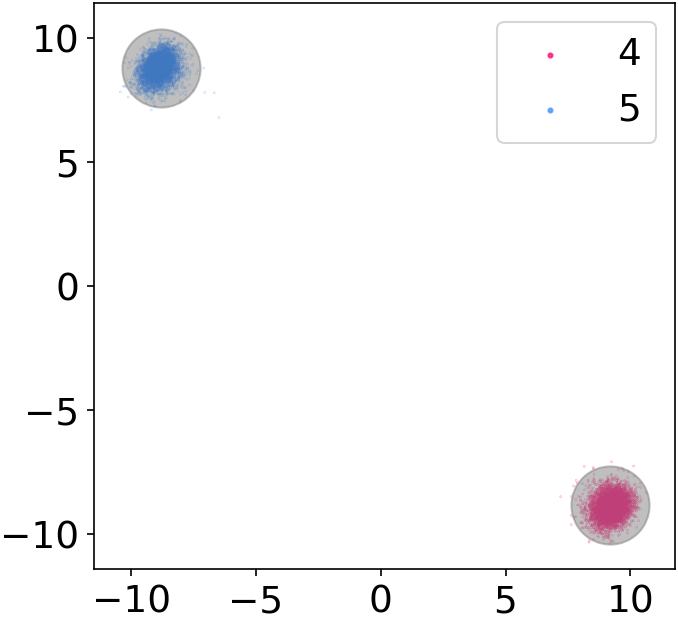} \\
    \rotatebox{90}{\hspace{1.5em} \textsc{Open set}} &
    \includegraphics[height=8em]{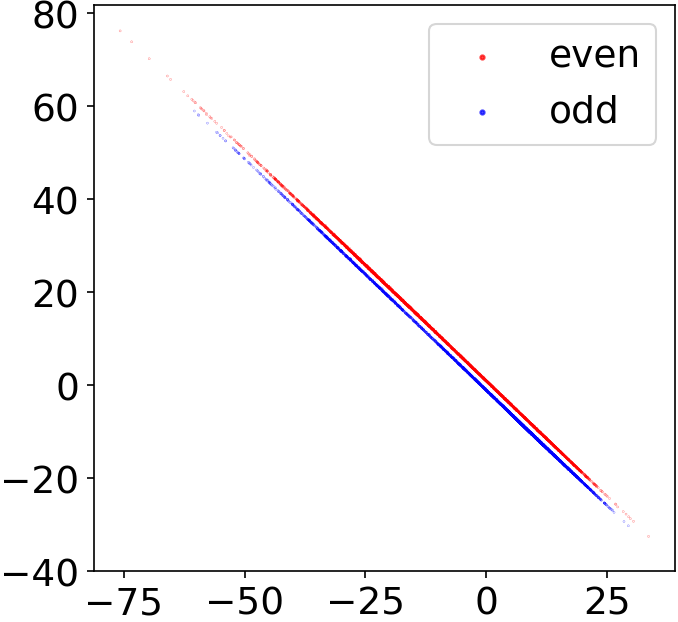} &
    \includegraphics[height=8em]{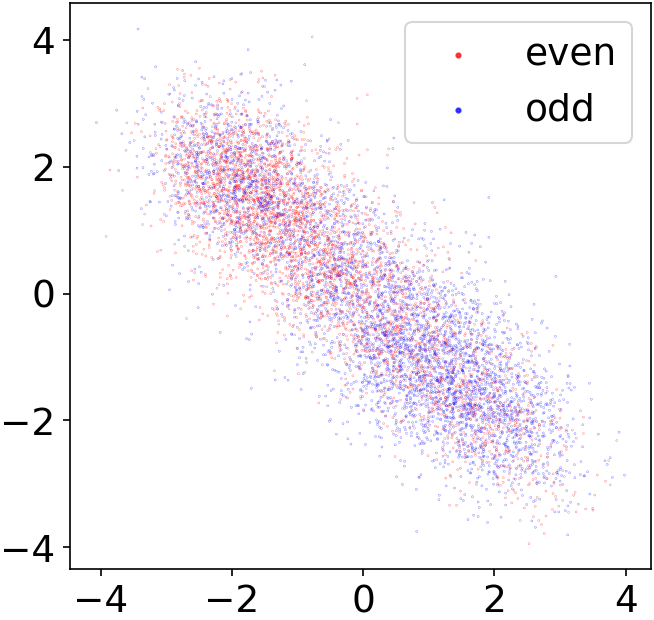} &
    \includegraphics[height=8em]{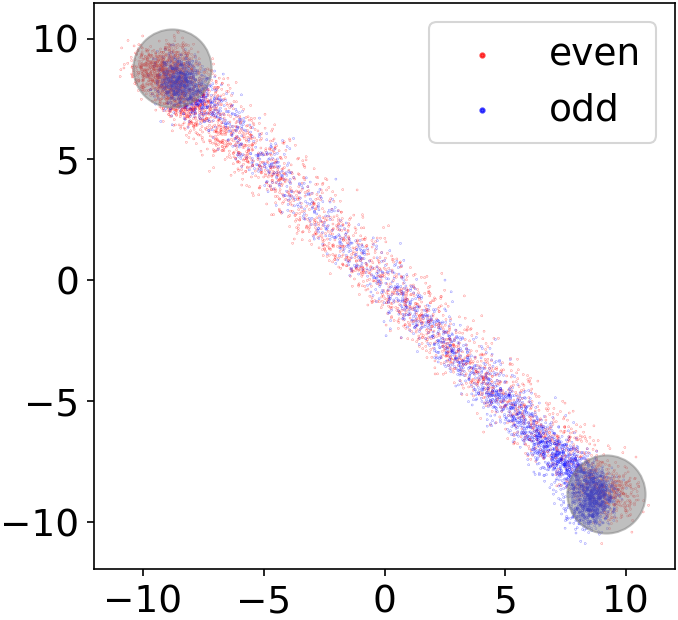}
    \end{tabular}
    
    \caption{\textbf{Latent space representation for MNIST.} On the first row, we report the representations for \textcolor{glancecolor}{4} and \textcolor{cbnmcolor}{5} as fitted by CBNM, VAE and \method, respectively. On the second row, we display the scattering plot for points only belonging to the open set. For CBNM, we separated \textcolor{red}{even} and \textcolor{blue}{odd} instances by $\Delta y=2$, since their representations strongly overlap. All plots comprise only the $z_4, z_5$ axes.  }
    \label{fig:app-representations-mnist}
\end{figure}

\subsection{Qualitative results}

\begin{figure}[!h]
    \centering
    \begin{tabular}{c||c}
    \includegraphics[height=7em]{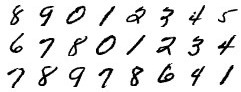} & 
    \includegraphics[height=7em]{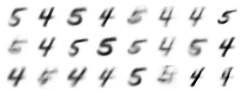}
    \end{tabular}
    
    \caption{\textbf{MNIST reconstruction with \method.} On the left we reported the original digits, whereas on the right the reconstruction with the learned decoder. All images have been inverted in the black and white scale.}
    \label{fig:recon-mnist}
\end{figure}

In~\cref{fig:app-representations-mnist}, we show the latent space representations for different models on the MNIST leakage test, for both closed-set and open-set data points. To illustrate the contribution of our mixture prior, in addition to the CBNM and \method models, we also considered a simpler supervised VAE model.  This model has the same encoder, decoder, and classifier as the \method, but uses a regular Gaussian prior\footnote{For the VAE model, we chose the Gaussian prior in \cite{kingma2014auto}, \ie $p(\vz) = \mathcal{N} (\vz | 0, 1)$. }. We found that this model achieved a similar level of leakage to CG-VAE. We display in~\cref{fig:recon-mnist} the reconstruction of a few random examples output by \method: the reconstructions of all instances belonging to the open-set greatly deviate from the original.

\section{Concept Leakage on dSprites} \label{sec: concept leakage dsprites}

{
In this section, we report additional material for the dSprites concept leakage experiment. This experiment resembles the previous one on MNIST:
\begin{enumerate}
    \item At training time, a CBM learns the representations of \texttt{shape}, \texttt{size} and \texttt{rotation} by receiving supervision on all possible variations of these factors. On the other hand, no variation of factors \texttt{pos}$\_$x and \texttt{pos}$\_$y are observed, in the sense that the position of the training sprites is fixed to the center of the image. We fit the CBNM and the \method with 5 latent factors to learn the representations. Again, no downstream classification task appears at this stage.
    \item At test time, the encoder is kept fixed for different variations varying  of the factors \texttt{pos}$\_$x and \texttt{pos}$\_$y are observed. The downstream task in this phase amounts to recognizing whether a sprite lies in top-right or in the bottom-left corner of the image.
\end{enumerate}
In this experiment, concept leakage occurs if the accuracy on the downstream task is above the $50\%$.
}

\begin{figure}[!h]
    \centering
    \begin{tabular}{cc}
    \includegraphics[height=15em]{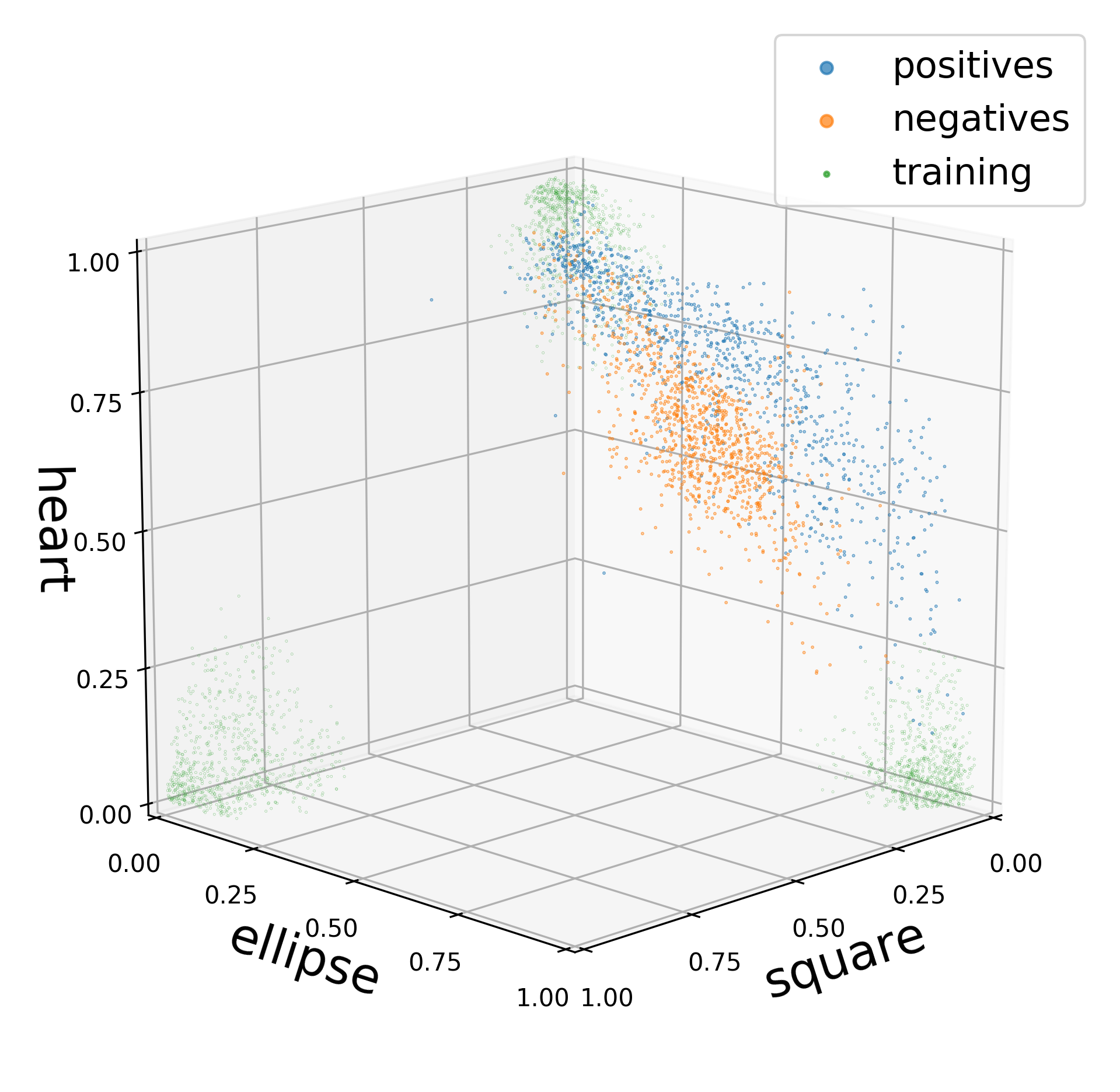} & 
    \includegraphics[height=13em]{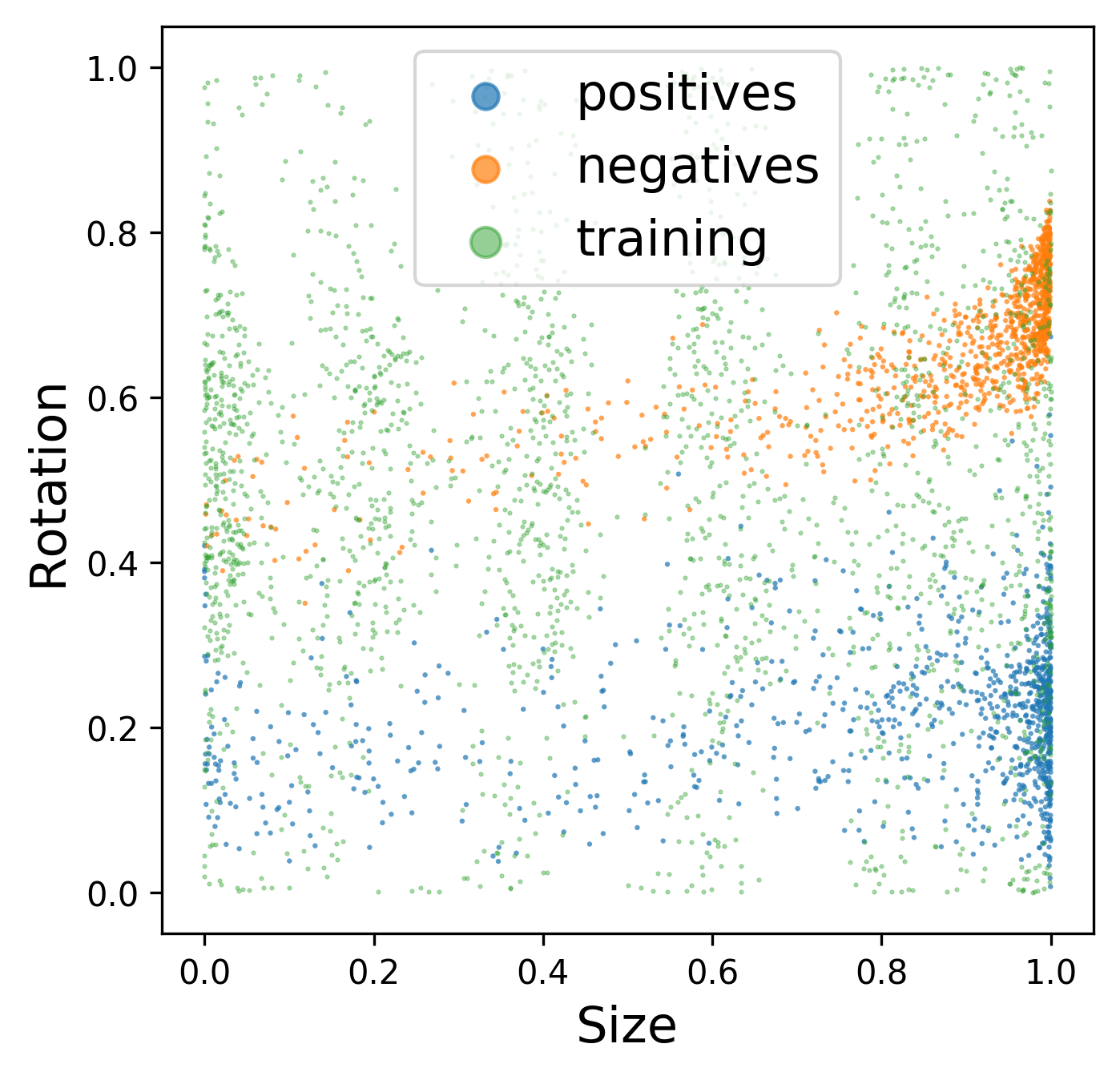}
    \end{tabular}
    
    \caption{\textbf{Concept space representation of \method for dSprites.}  On the left, we show the projections on the one-hot encoded \texttt{shape} subspace, whereas on the right we project on the \{\texttt{size}, \texttt{rotation}\} subspace. We include the representations for \textcolor{ForestGreen}{training} points, \textcolor{blue}{positive} and \textcolor{orange}{negative} ones. }
    \label{fig:repr-glance-dsprites}
\end{figure}

\subsection{Qualitative results}

\begin{figure}[!h]
    \centering
    \begin{tabular}{cc}
    \footnotesize \rotatebox{90}{\hspace{1em} \textsc{Closed set}} &
    \includegraphics[height=7em]{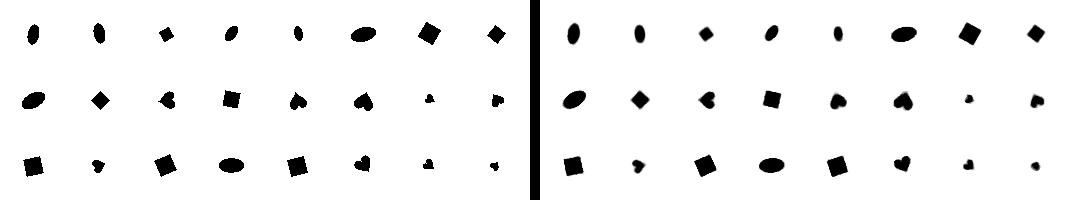} \\
    \hline
    \footnotesize \rotatebox{90}{\hspace{1em} \textsc{Open set}} & \includegraphics[height=7em]{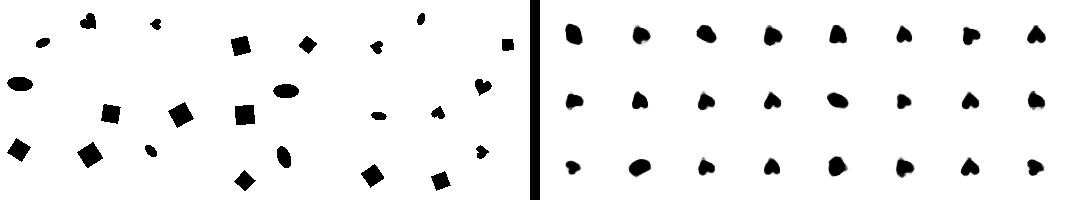}
    \end{tabular}
    
    \caption{\textbf{Reconstruction for dSprites on train and test with \method.} On the upper panel, we report the reconstructions of the sprites belonging to the closed set. On the lower one, the reconstructions of the open set points. Like MNIST, all images have been inverted in the black and white scale.}
    \label{fig:reco-dsprites}
\end{figure}

We also include qualitative results for \method and  on dSprites for closed set and open set data points.
In~\cref{fig:repr-glance-dsprites} we display the projections of train and test points on the two different latent subspaces (see caption). In both of them, \textcolor{blue}{positives} and \textcolor{orange}{negatives} representations are well separated from each other, implying substantia leakage.
We also evaluated the reconstruction quality during training and testing and reported some of them in~\cref{fig:reco-dsprites}. Notably, almost all points are recognized to be open set instances thanks to the reconstruction threshold.

{
\section{Additional results for \methods and CBNMs in CelebA}

In this section, we discuss additional results for CBNMs vs.\@ \methods on the CelebA dataset. We first report the accuracy of the learned concepts on the supervised latent factors for both CBNMs and \methods in CelebA. Then, we examine two variants of \methods varying the dimension of the unsupervised factors in the latent space: a $\beta$-VAE with 20 latent factors and a $\beta$-TCVAE with 40 latent factors, \cite{chen2018isolating}. This variant includes an additional loss term 
given by the Total Correlation (TC) of the model posterior $q_\phi(\vz) = \bbE_{p_\mathcal{D} (\vx)} [q_\phi(\vz | \vx) ]$: 
\begin{equation}
    (\beta - 1) \cdot \mathsf{KL}\big(q_\phi(\vz) || \prod_{i=1}^k q_\phi(z_i) \big)
\end{equation}
where $\beta$ denotes the strength hyper-parameter. Both the $\beta$-VAE and the $\beta$-TCVAE receive supervision only on the 10 generative factors that are fitted in the CBNM. A the end of the section, we report traversals for the models with $40$ latent factors.

\subsection{Concepts Accuracy}

We report the concepts accuracy for both CBNMs and \methods in \cref{fig:concept accuracy}, with 10 latent dimensions and the TCVAE variant. The difference in concept accuracy between \method (both variants) and CBNMs is negligible, with \methods showing slightly higher variance when the percentage of concepts supervision is very small.  This highlights how, in terms of concept accuracy, the two classes of models are essentially indistinguishable, even though they are in terms of alignment.

\begin{figure}[!h]
    \centering
    \begin{tabular}{cc}
    \textsc{\hspace{0.3em} CBNMS $vs$ \methods(10)} & \textsc{\hspace{0.3em} CBNMS $vs$ \methods(40)} \\
\includegraphics[width=0.49\textwidth]{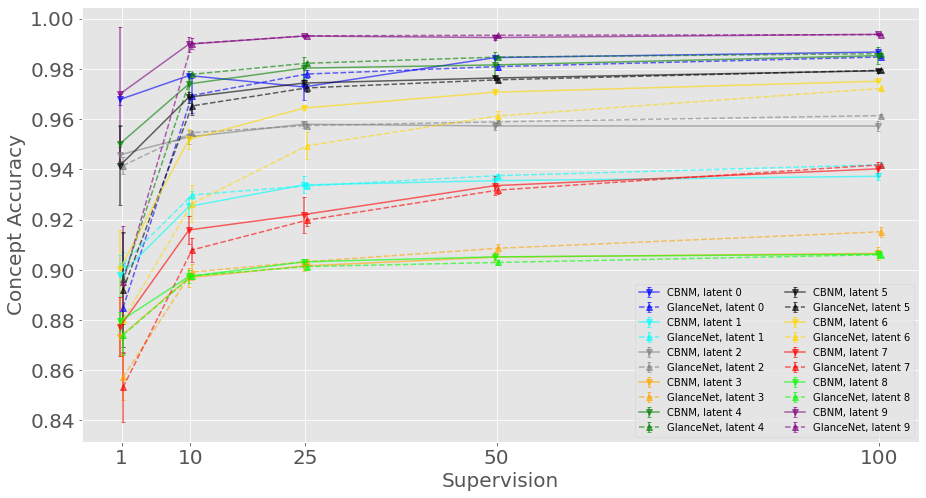}
         &  
 \includegraphics[width=0.49\textwidth]{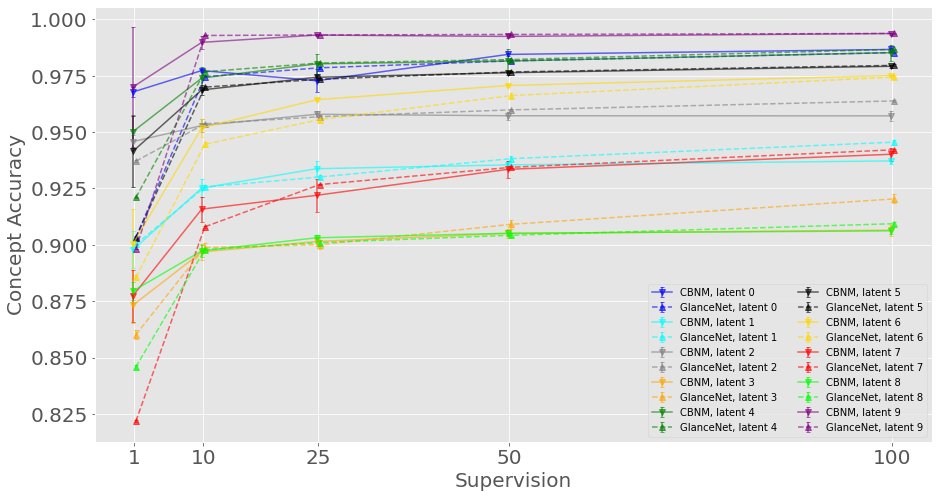}
    \end{tabular}
    
    \caption{Concepts accuracy for CBNMs $vs$ \methods. Different colors refer to the distinct attributes for which supervision is provided. The solid line is reserved to CBNMs, whether \methods are displayed with a dotted line. On the left, CBNMs vs \methods with a latent space of dimension 10. On the right, CBNMs vs \methods with TCVAE variant and a latent space of 40.  }
    \label{fig:concept accuracy}
\end{figure}

\subsection{Performances upon variations of the latent space dimension}

Here, we show the behavior of the metrics upon increasing the dimension of the latent space. The first variant of \methods, based on $\beta$-VAE, was fitted with $\beta\approx1$, with a latent space of dimension 20. The second variant is a TCVAE, trained with a weight of the total correlation $\beta=10$ for all concepts supervisions, exception made for the $100$\% run, where we found better results with $\beta=0.5$. 
We measured alignment and explicitness for both variants of \methods by restricting on only those 10 latent factors where supervision were provided. This is in line with the notion of alignment, since we are interested in measuring the interpretability of the model, not the disentanglment among different components.  

In \cref{fig: celeba comparison different glance} we report the results obtained, including the original variant with 10 latent dimensions. For the $\beta$-VAE (20) and TCVAE (40) we can see the improvement provided by extending the latent space. The latter achieves particular high values of alignment.

\begin{figure}[!h]
    \centering
    \begin{tabular}{cccc}
        & \textsc{Accuracy} & \textsc{Alignment} & \textsc{Explicitness} \\   
        \rotatebox{90}{\hspace{0.2em} \textsc{ $\beta$-VAE (10) } } & \includegraphics[width=0.275\linewidth]{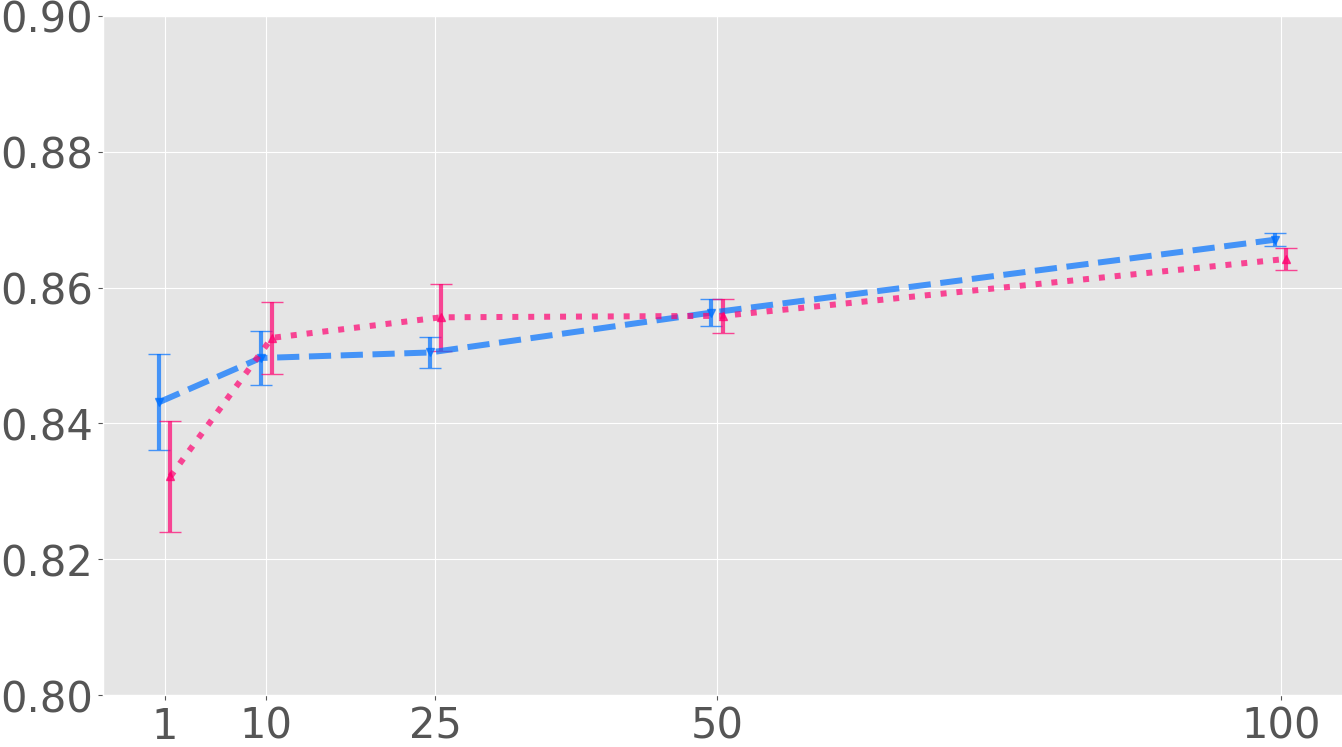} & 
        \includegraphics[width=0.275\linewidth]{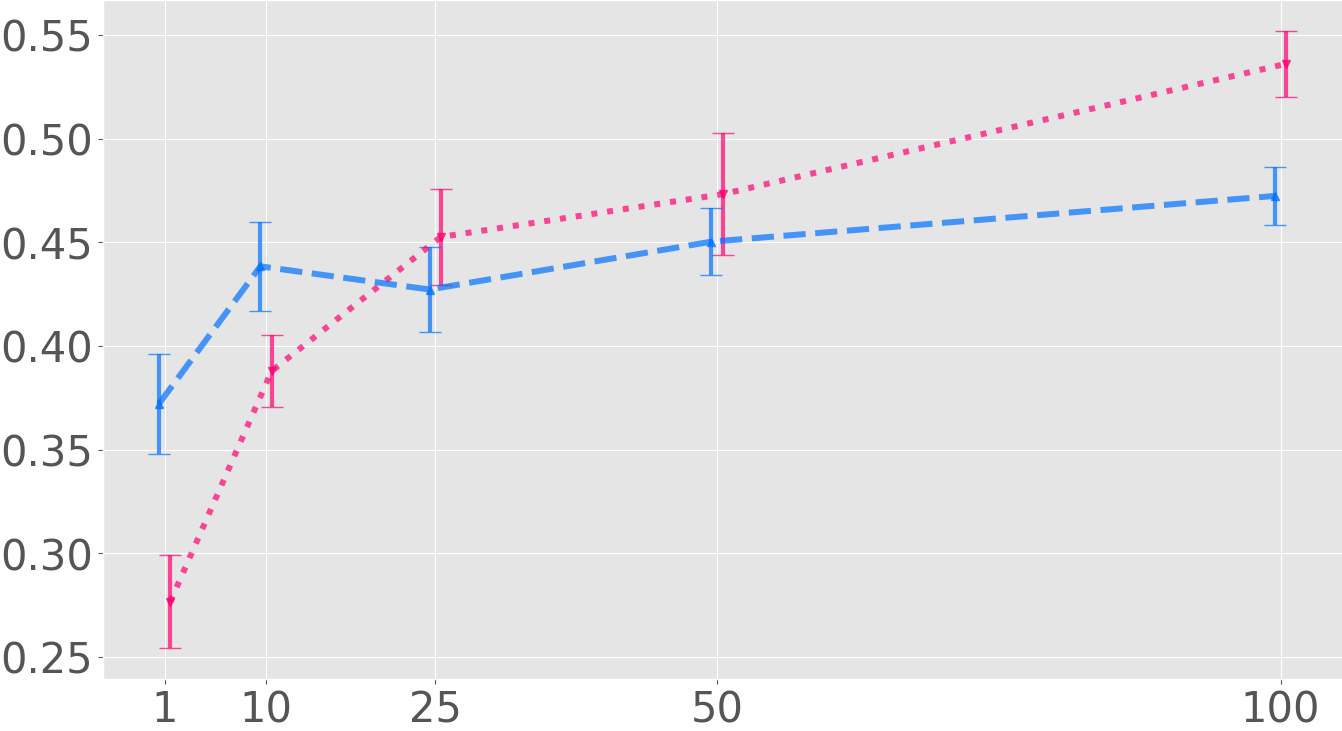} &
        \includegraphics[width=0.275\linewidth]{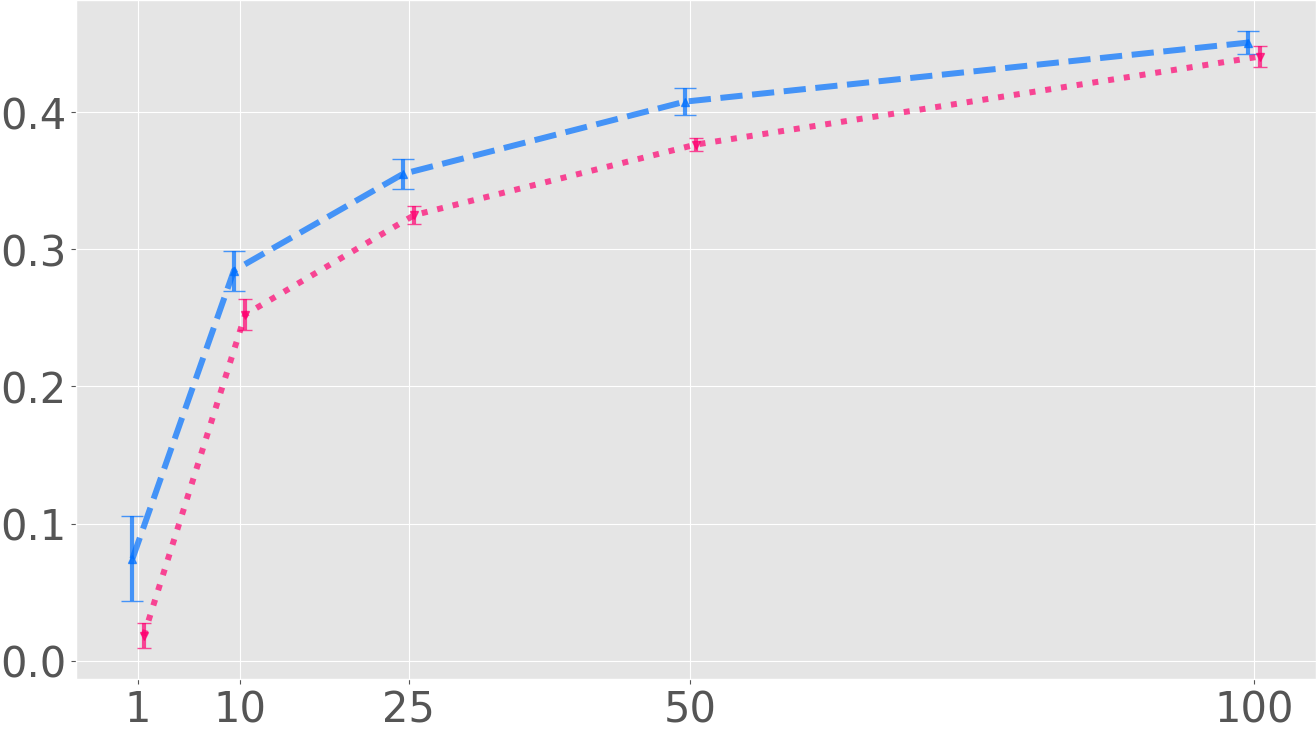}
        \\
        \rotatebox{90}{\hspace{0.2em} \textsc{ $\beta$-VAE (20) } } &  
        \includegraphics[width=0.275\linewidth]{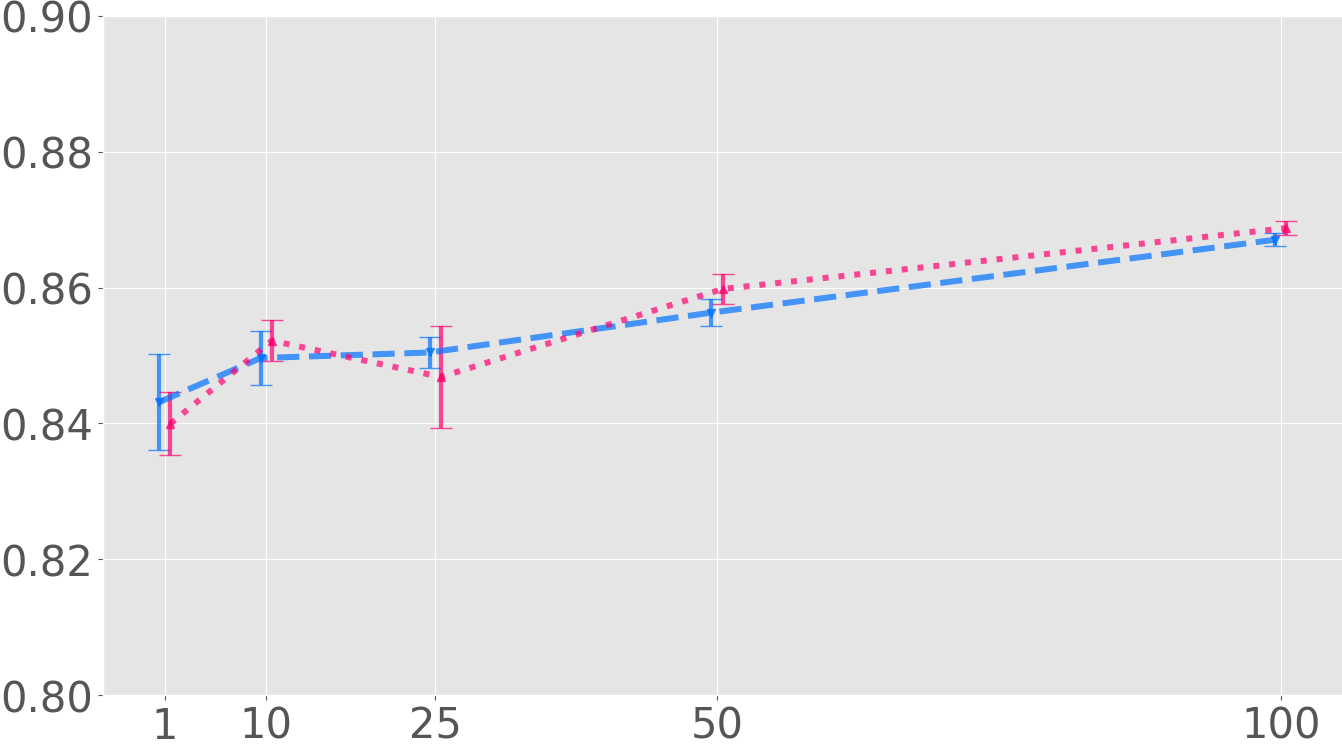} & 
        \includegraphics[width=0.275\linewidth]{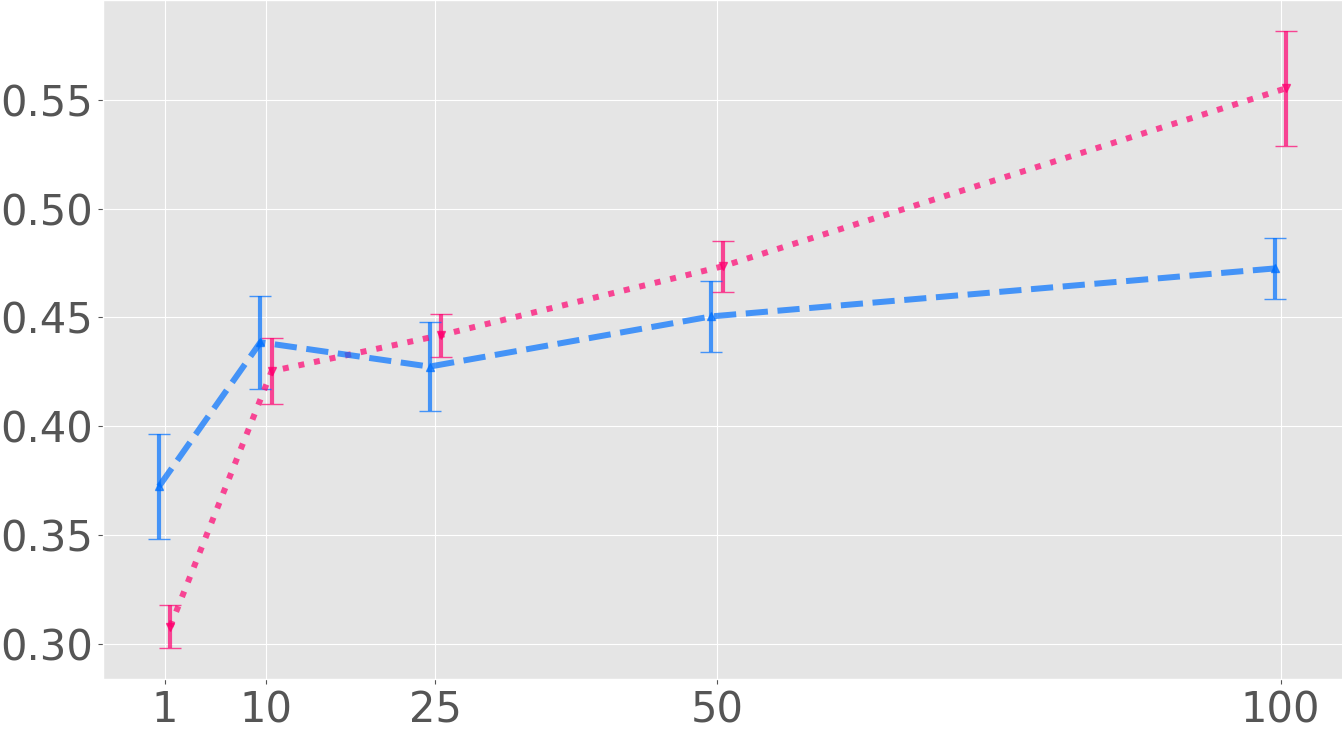} &
        \includegraphics[width=0.275\linewidth]{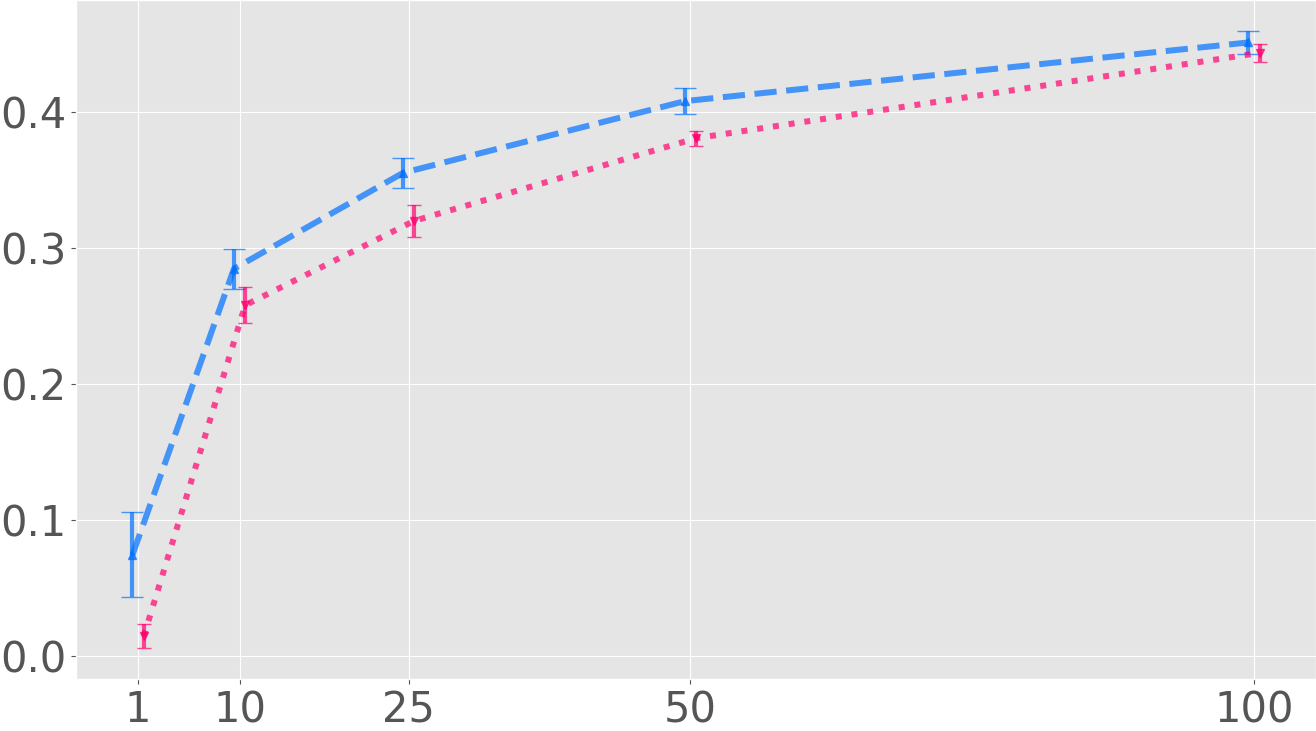}\\
        \rotatebox{90}{\hspace{0.2em} \textsc{ TCVAE (40) } } &  
        \includegraphics[width=0.275\linewidth]{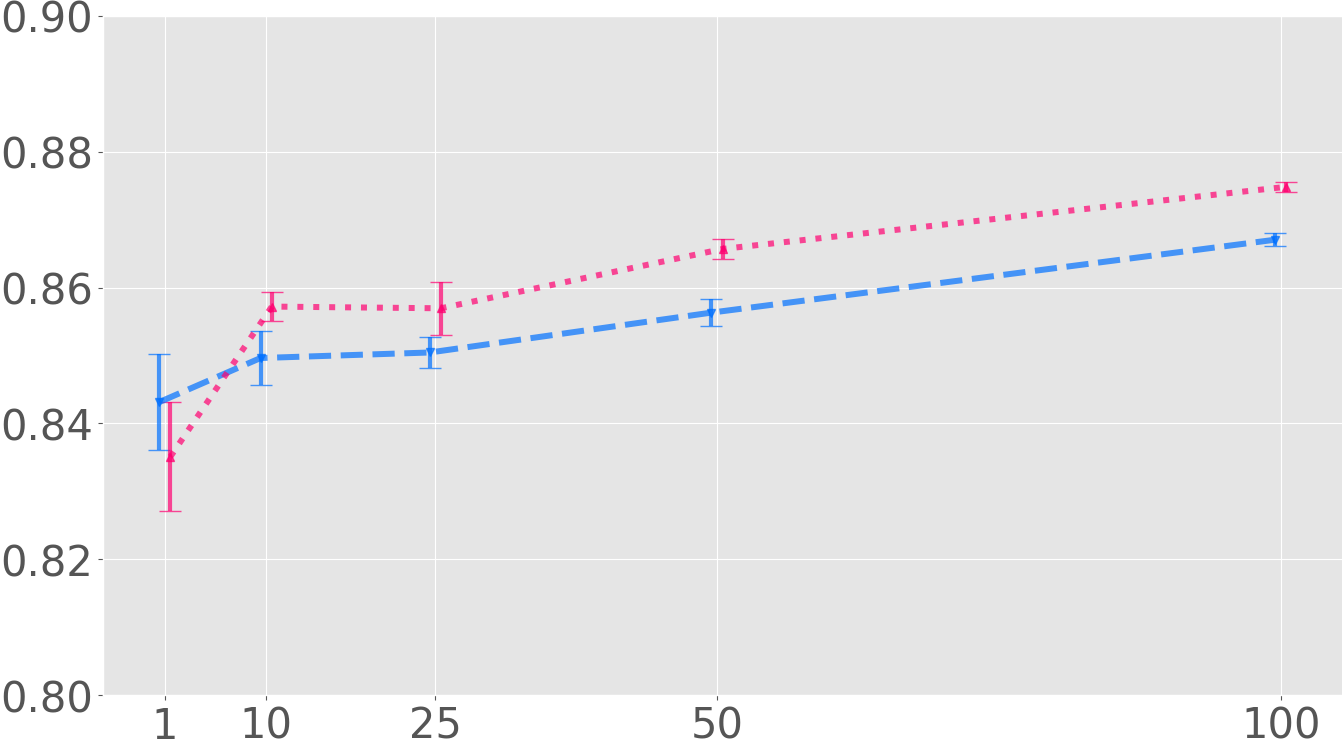} & 
        \includegraphics[width=0.275\linewidth]{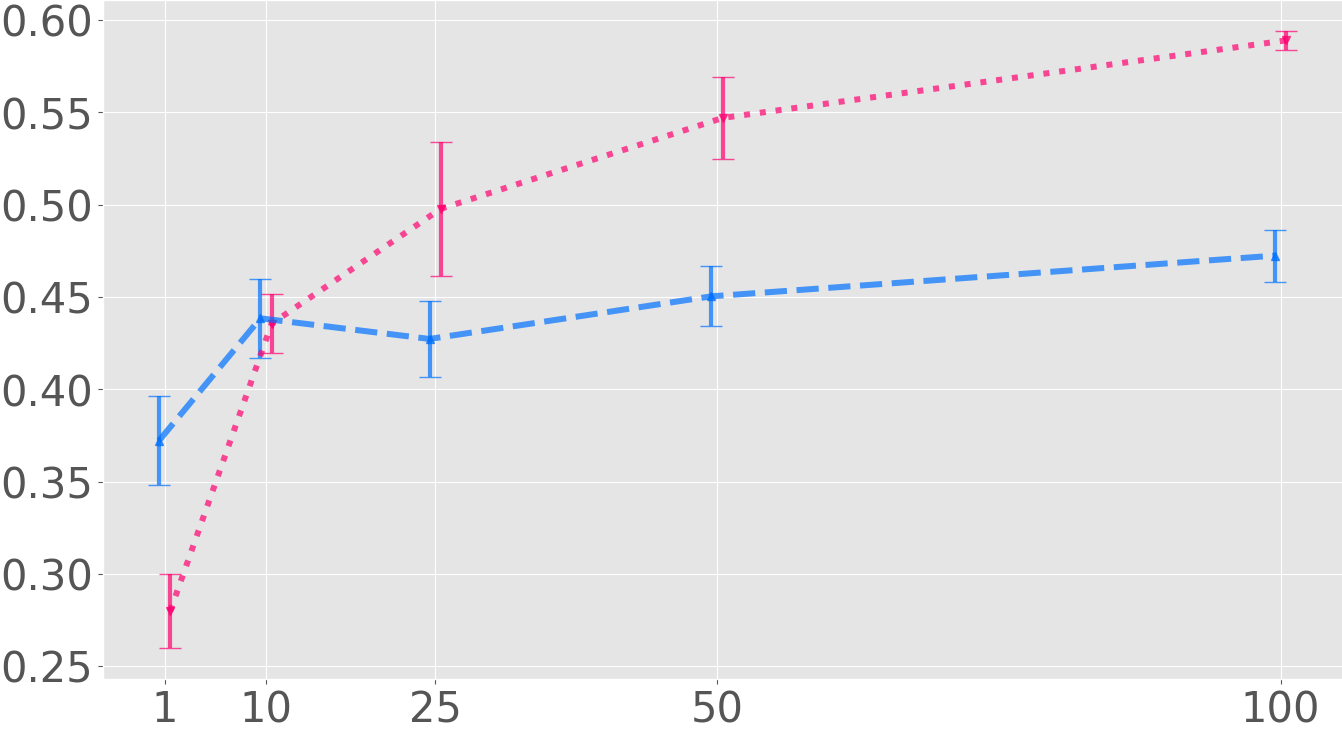} &
        \includegraphics[width=0.275\linewidth]{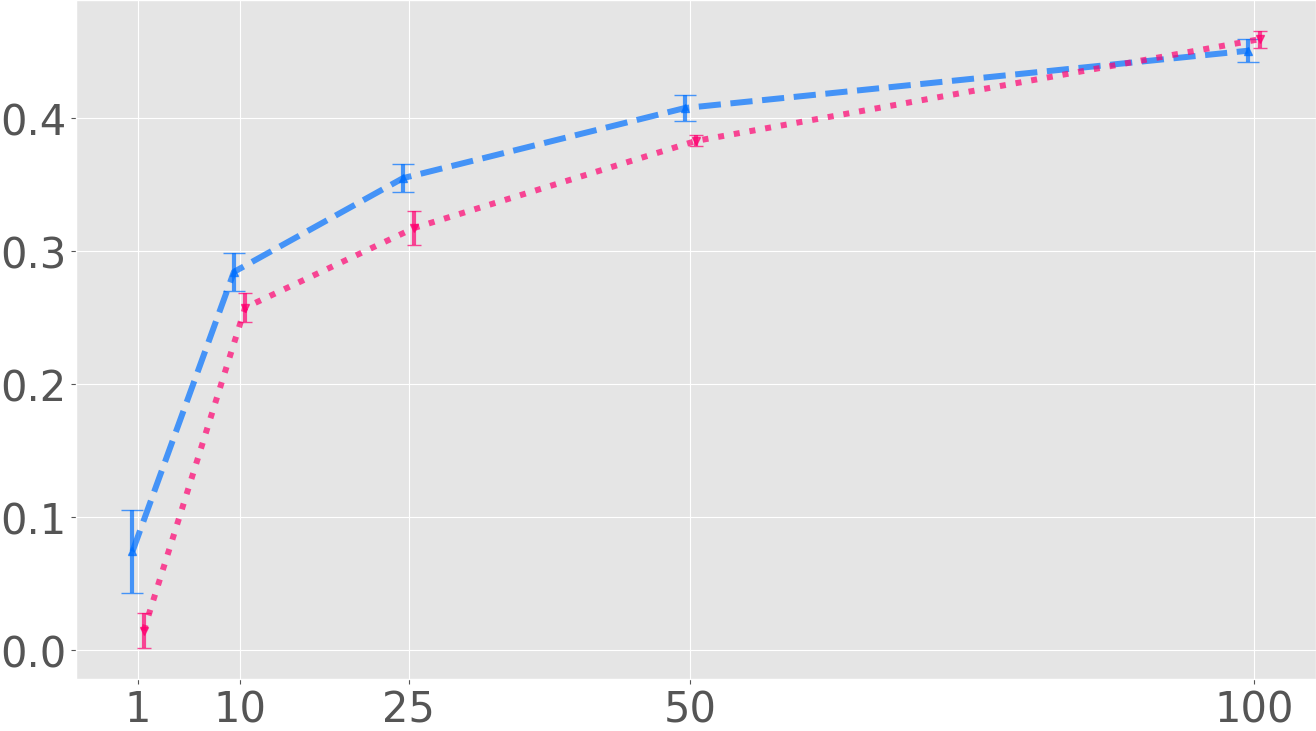}
    \end{tabular}
    \caption{Accuracy, Alignment and Explicitness metrics for CBNMs $vs$ \methods. For each row we vary the comparison with variants of \methods: $\beta$-VAE (10) refers to the model we reported in the main text, $\beta$-VAE (20) is a variant with 20 latent dimensions, and TCVAE (40) is the model based on a TCVAE with 40 latent dimensions.} 
    \label{fig: celeba comparison different glance}
\end{figure}

\subsection{Latent traversals}
We finally report in  the traversals for some of the supervised attributes, obtained by the \method TCVAE with full supervision on the concepts. We excluded the traversals of the attributes \textsc{hat} and \textsc{bald} since the generator failed to reproduce them faithfully. The others are well captured by the model, as we reported in \cref{fig: latent traversal}.

\begin{figure}[!h]
    \centering
    \includegraphics[width=\textwidth]{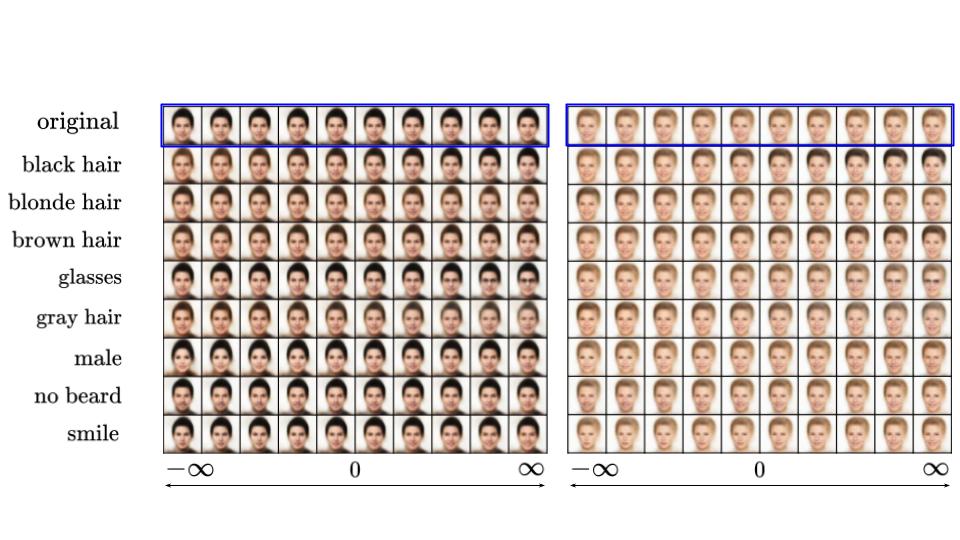}
    \caption{Latent traversals on two test images.  In each row, we report the result of changing a single latent factor $Z_i$ (from $-5$ to $+5$) while keeping fixed the others.}
    \label{fig: latent traversal}
\end{figure}
}



\end{document}